\documentclass[journal]{IEEEtran}
\ifCLASSINFOpdf
\else
\fi
\usepackage{times}  
\usepackage{helvet} 
\usepackage{courier}  
\usepackage[hyphens]{url}  
\usepackage{graphicx} 
\usepackage{caption} 

\usepackage{latexsym}
\usepackage{algorithm}
\usepackage{algorithmic}   
\usepackage{amsmath}
\usepackage{amsfonts}
\usepackage{url}
\usepackage[switch]{lineno}
\usepackage{booktabs}
\usepackage{threeparttable}  
\usepackage{graphicx}
\usepackage{subfigure}
\usepackage{multirow}
\usepackage{amssymb}
\usepackage{placeins}
\usepackage{float}
\usepackage{color}
\usepackage{verbatim}
\usepackage{bm}
\usepackage{hyperref}
\usepackage{adjustbox}
\usepackage{array}
\usepackage{tabularx}
\usepackage{epstopdf}
\begin{document}
%
\title{HyperTTA: Test-Time Adaptation for Hyperspectral Image Classification under Distribution Shifts
}
%
%
%

\author{Xia Yue,
	Anfeng Liu,
        Ning Chen,
        Chenjia~Huang,
        Hui~Liu,
        Zhou~Huang,
        and Leyuan Fang,~\IEEEmembership{Senior~Member,~IEEE}
	\thanks{This work was supported in part by the National Natural Science Foundation of China under Grant U2344216, Grant 62425109 and Grant U22B2014; in part by the Science and Technology Plan Project Fund of Hunan Province under Grant 2022RSC3064.}
        \thanks{Xia Yue is with School of Computer Science and Engineering, Central South University, Changsha 410083, China. (e-mail: yuexia486@gmail.com).}
        \thanks{Anfeng Liu is with the School of Electronic Information, Central South University, Changsha 410083, China. (e-mail: afengliu@mail.csu.edu.cn).}

        \thanks{Ning Chen and Zhou Huang are with the Institute of Remote Sensing and Geographic Information System, School of Earth and Space Sciences, Peking University, Beijing 100871, China (e-mail: chenning0115@pku.edu.cn).}
	
     \thanks{Chenjia Huang and Hui Liu are with the School of Remote Sensing \& Geomatics Engineering, Nanjing University of Information Science \& Technology, Nanjing 210044, China (email: cjhuang@nuist.edu.cn; huil@pku.edu.cn).}%
    
    \thanks{Leyuan Fang is with the College of Electrical and Information Engineering, Hunan University, Changsha 410082, China, and also with the Peng Cheng Laboratory, Shenzhen 518000, China (e-mail: fangleyuan@gmail.com).}%
	\thanks{}}

%
%

\markboth{}%
{Shell \MakeLowercase{\textit{et al.}}: Bare Demo of IEEEtran.cls for IEEE Journals}
%



\maketitle

\begin{abstract}
Hyperspectral image (HSI) classification models are highly sensitive to distribution shifts caused by real-world degradations such as noise, blur, compression, and atmospheric effects. To address this challenge, we propose HyperTTA (Test-Time Adaptable Transformer for Hyperspectral Degradation), a unified framework that enhances model robustness under diverse degradation conditions. First, we construct a multi-degradation hyperspectral benchmark that systematically simulates nine representative degradations, enabling comprehensive evaluation of robust classification. Based on this benchmark, we develop a Spectral–Spatial Transformer Classifier (SSTC) with a multi-level receptive field mechanism and label smoothing regularization to capture multi-scale spatial context and improve generalization. Furthermore, we introduce a lightweight test-time adaptation strategy, the Confidence-aware Entropy-minimized LayerNorm Adapter (CELA), which dynamically updates only the affine parameters of LayerNorm layers by minimizing prediction entropy on high-confidence unlabeled target samples. This strategy ensures reliable adaptation without access to source data or target labels. Experiments on two benchmark datasets demonstrate that HyperTTA outperforms state-of-the-art baselines across a wide range of degradation scenarios. The code will be made publicly available at https://github.com/halfcoder1/HyperTTA.
\end{abstract}

\begin{IEEEkeywords}
Hyperspectral image classification; Distribution shift; Multi-degradation benchmark; Test-time adaptation; Transformer
\end{IEEEkeywords}

%
\IEEEpeerreviewmaketitle

\section{INTRODUCTION}
\IEEEPARstart{H}{yperspectral} remote sensing has gained significant attention in recent years due to its ability to capture dense spectral information across a continuous range of wavelengths for each pixel~\cite{Huang2022}. This rich spectral content supports diverse applications such as environmental monitoring~\cite{rajabi2024editorial}, precision agriculture~\cite{foerster2024hyperedu}, mineral exploration~\cite{hajaj2024review}, and urban planning~\cite{deluca2024improving}, enabling the identification of subtle surface materials that are often indistinguishable in conventional multispectral imagery.

Despite its potential, HSI classification remains challenging due to the high dimensionality of the data—commonly referred to as the “curse of dimensionality”~\cite{song2024study}—and the scarcity of labeled training samples~\cite{10632198}. Additionally, strong intra-class variability and inter-class similarity~\cite{10554657,10659915,10816304}, along with environmental and acquisition-related factors such as illumination shifts, atmospheric conditions, and sensor noise~\cite{10475370,10440631}, introduce significant distribution shifts between training and testing domains, further complicating the classification task.

\begin{figure}[t]
\centering
\includegraphics[width=0.5\textwidth]{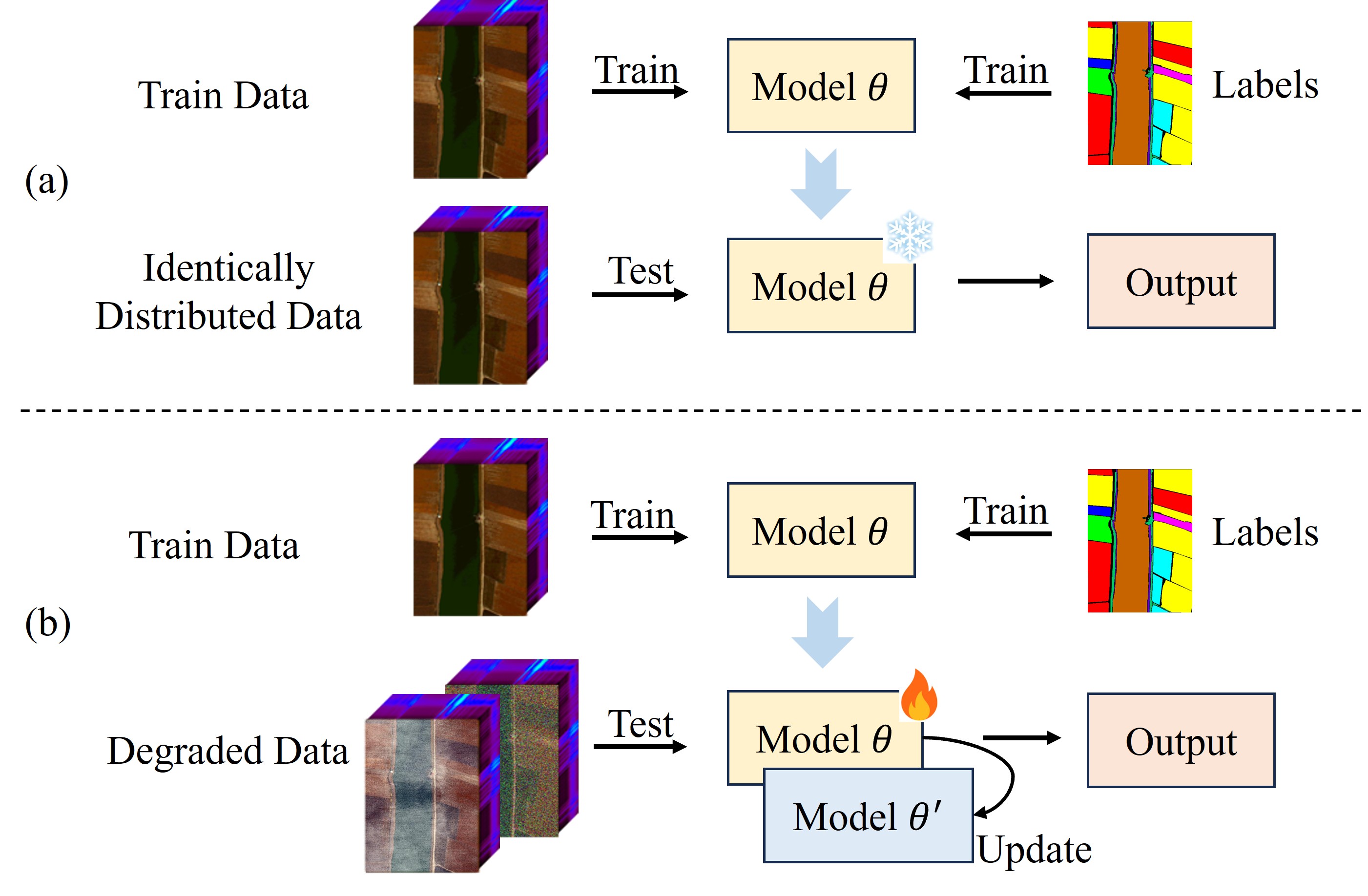}
\caption{Difference between traditional and TTA method. (a): Traditional method; (b): TTA method.}
\label{fig:TTA_diff}
\end{figure}

To cope with these challenges, TTA has emerged as a promising paradigm~\cite{10420496,Liang2024}. In contrast to traditional domain adaptation methods that require target-domain access during training~\cite{adp6040,9944086,WANG2018135}, TTA enables model refinement at inference time using unlabeled test data, typically via entropy minimization~\cite{Wang2024,yuan2023robust}, batch normalization calibration, or self-ensembling techniques. This flexibility makes TTA particularly suitable for real-world remote sensing deployments, where dynamic environmental changes and limited target annotations are common. Fig.~\ref{fig:TTA_diff} illustrates the difference between traditional and TTA method, the (a) shows traditional method and (b) shows TTA method. 

Although TTA has achieved success in fields such as natural image classification and medical imaging~\cite{chen2022contrastive,wang2020tent,ma2024swapprompt,valanarasu24a,wen2024denoising}, its application to hyperspectral remote sensing is still underexplored. Existing methods largely assume stable test distributions and operate on low-dimensional RGB inputs, which do not reflect the complex spectral characteristics of HSI data~\cite{9645266}. Furthermore, traditional domain adaptation approaches often rely on multi-stage retraining or access to labeled target samples~\cite{10.1145/3631712,10.1007/978-3-030-71704-9_65}, limiting their practicality. In addition, there is a lack of benchmark datasets and standardized evaluation protocols for TTA under realistic HSI degradations. These gaps highlight the urgent need for efficient and scalable TTA strategies tailored to hyperspectral data.

In this paper, we propose a novel TTA-based framework for robust HSI classification under diverse degradation conditions. Our main contributions are summarized as follows:
\begin{itemize}
    \item We construct and release the first hyperspectral TTA benchmark that systematically simulates nine types of real-world degradations, including fog, stripe noise, Gaussian noise (additive and zero-mean), Poisson noise, convolutional blur, salt-and-pepper noise, JPEG compression, and deadline noise. This dataset enables standardized evaluation of model robustness and adaptation under diverse degradation scenarios.
    \item We design SSTC Module that integrates a multi-level receptive field mechanism to capture spatial features at multiple scales, along with a label smoothing strategy to improve generalization under noisy supervision. This model enhances spectral–spatial representation learning and robustness to input corruptions.
    \item We propose a confidence-aware TTA method called CELA Module, which updates only the affine parameters of Layer Normalization layers during inference by minimizing prediction entropy on high-confidence target samples. This robust and source-free adaptation strategy improves generalization to unseen target distributions without requiring source data or target labels.
    \item By integrating the SSTC, and CELA modules, we construct a unified TTA framework named \textbf{HyperTTA}, tailored for HSI classification under severe degradations. Extensive experiments on PU and WHLK datasets demonstrate that HyperTTA outperforms existing classification and TTA baselines in terms of accuracy, robustness, and generalization ability across diverse corruption types.
\end{itemize}

\section{Related Work}
\subsection{Deep Learning for HSI Classification}
Deep learning approaches have attracted increasing attention in hyperspectral remote sensing due to their ability to automatically extract representative spectral–spatial features from raw data~\cite{Tejasree2024,PAOLETTI2019279}. Convolutional Neural Networks (CNNs) have been extended to handle high-dimensional inputs through variants such as 3D-CNNs, which jointly model spectral and spatial information~\cite{10398446}. To improve efficiency while maintaining accuracy, Tri-CNN~\cite{rs15020316} introduced a multi-scale 3D-CNN architecture with three-branch feature fusion, enabling better classification with reduced computational cost.

Recently, Transformer-based models have demonstrated superior performance in HSI classification by leveraging self-attention mechanisms to capture long-range dependencies and complex spectral correlations~\cite{rs13030498,9766028,9627165}. For example,~\cite{10604879} introduced a Transformer with implicit conditional positional encoding to handle variable-length inputs, while~\cite{SHU2024107351} proposed a dual-attention structure that captures both global spatial–spectral context and local spectral features. To better integrate multiscale spatial–spectral information,~\cite{10772125} applied random masking and token-aware pooling. Hybrid models that combine CNN and ViT, such as~\cite{10433662}, have also been proposed to bridge local and global feature modeling. These Transformer architectures have significantly advanced the field, demonstrating strong generalization across complex scenes~\cite{rs15071860,10399798}.

Despite these advances, deep learning-based HSI classifiers typically require large amounts of labeled data~\cite{Alzubaidi2023}, which are expensive and labor-intensive to obtain~\cite{9862940}. Moreover, models trained on one dataset (source domain) often suffer severe performance drops when applied to data with different distributions (target domain)~\cite{chen2023improved}. These challenges underscore the need for more robust methods that can maintain high classification accuracy under label scarcity and distribution shifts—especially under diverse real-world degradations, which are seldom considered in most existing work.

\subsection{Robust Learning under Degradation}
In real-world remote sensing applications, HSIs are frequently affected by various types of degradations caused by sensor noise, atmospheric interference, compression artifacts, or optical distortions~\cite{9552462}. To address this issue, several studies have attempted to improve model robustness by injecting synthetic noise during training or designing denoising pre-processing pipelines~\cite{10632198,9932642}. However, most existing works are limited to specific or isolated degradation types and do not reflect the complexity of real-world multi-source corruptions~\cite{9826537}. Furthermore, there is currently a lack of unified datasets or benchmarks that systematically simulate and evaluate multiple degradation scenarios in a controlled setting. This makes it difficult to develop or compare robust learning and adaptation methods under consistent evaluation protocols.

To fill this gap, we construct a comprehensive multi-degradation hyperspectral dataset that includes nine representative corruption types, covering noise (e.g., Gaussian, Poisson, Salt-and-Pepper), structural degradation (e.g., stripes, deadlines), blur, compression, and fog. This dataset enables systematic evaluation of classification and adaptation models under diverse and challenging degradation conditions, and serves as a foundation for our proposed robust learning framework.

\subsection{Domain Adaptation in HSI Classification}
Domain adaptation (DA) techniques aim to mitigate distribution shifts between training (source) and testing (target) domains~\cite{10452835,FANG2024106230}. In the hyperspectral domain, various strategies have been explored to align feature distributions across domains, including adversarial learning~\cite{10268909}, manifold alignment~\cite{10520907}, and transfer learning~\cite{9775658}. DA methods can be categorized into supervised, semi-supervised, and unsupervised approaches, depending on the availability of labeled target-domain samples~\cite{10.1007/978-3-030-71704-9_65}.

Recent advances have introduced sophisticated spectral–spatial DA frameworks. For example,~\cite{9924236} proposed a two-branch attention-based adversarial model to enhance cross-domain feature alignment, while~\cite{10440363} integrated adaptive sampling and adversarial optimization to handle class imbalance and complex scene variability. In another work,~\cite{10620320} applied masked self-distillation to improve feature discriminability in unsupervised domain adaptation settings.

Despite their effectiveness, DA methods typically require access to both source and target domain data during training~\cite{adp6040,9944086}. However, in practical hyperspectral remote sensing scenarios, target-domain data are often scarce, highly variable, or unavailable during training. This limits the applicability of traditional DA techniques and motivates alternative approaches such as TTA, which can adapt models dynamically without relying on target labels or pre-training on the target domain.

\subsection{TTA Concepts and Methods}

TTA has emerged as a flexible solution to address distribution shifts during inference, without requiring labeled target domain data.~\cite{schirmer2024testtime}. Unlike conventional domain adaptation methods that require access to source and target data during training~\cite{adp6040,9944086,WANG2018135}, TTA adapts the model online at test time, typically by updating normalization statistics~\cite{Wang2024}, minimizing prediction uncertainty~\cite{zhang2023domainadaptor}, or leveraging self-ensembling techniques~\cite{jia2024tinytta}. TTA has shown strong performance in fields such as natural image recognition and medical imaging, offering a practical balance between adaptation capability and annotation efficiency.

In the context of remote sensing, however, TTA remains underexplored. Most existing efforts primarily recalibrate batch normalization statistics on optical (RGB) remote sensing data~\cite{LIANG2025104463}, while little attention has been paid to hyperspectral data, which presents unique challenges such as high spectral dimensionality, strong reflectance variability, and domain shifts caused by atmospheric and temporal factors~\cite{10543066}. Moreover, the large volume of HSI data makes real-time adaptation computationally demanding. These factors collectively hinder the direct application of existing TTA methods to hyperspectral scenarios.

To address these limitations, this paper proposes a TTA framework tailored specifically for HSI classification. Unlike prior TTA approaches focused on natural or medical images, our method performs entropy minimization on unlabeled target data while updating only the affine parameters of Layer Normalization layers. This design maintains model compactness while enabling dynamic adaptation to unseen degradation patterns. Coupled with our multi-degradation hyperspectral dataset, the proposed method offers a practical and efficient solution for real-world remote sensing applications where source data may be inaccessible and target distributions are non-stationary.

\section{Method} 
\begin{figure*}[htbp]
\centering
\includegraphics[width=1.0\textwidth]{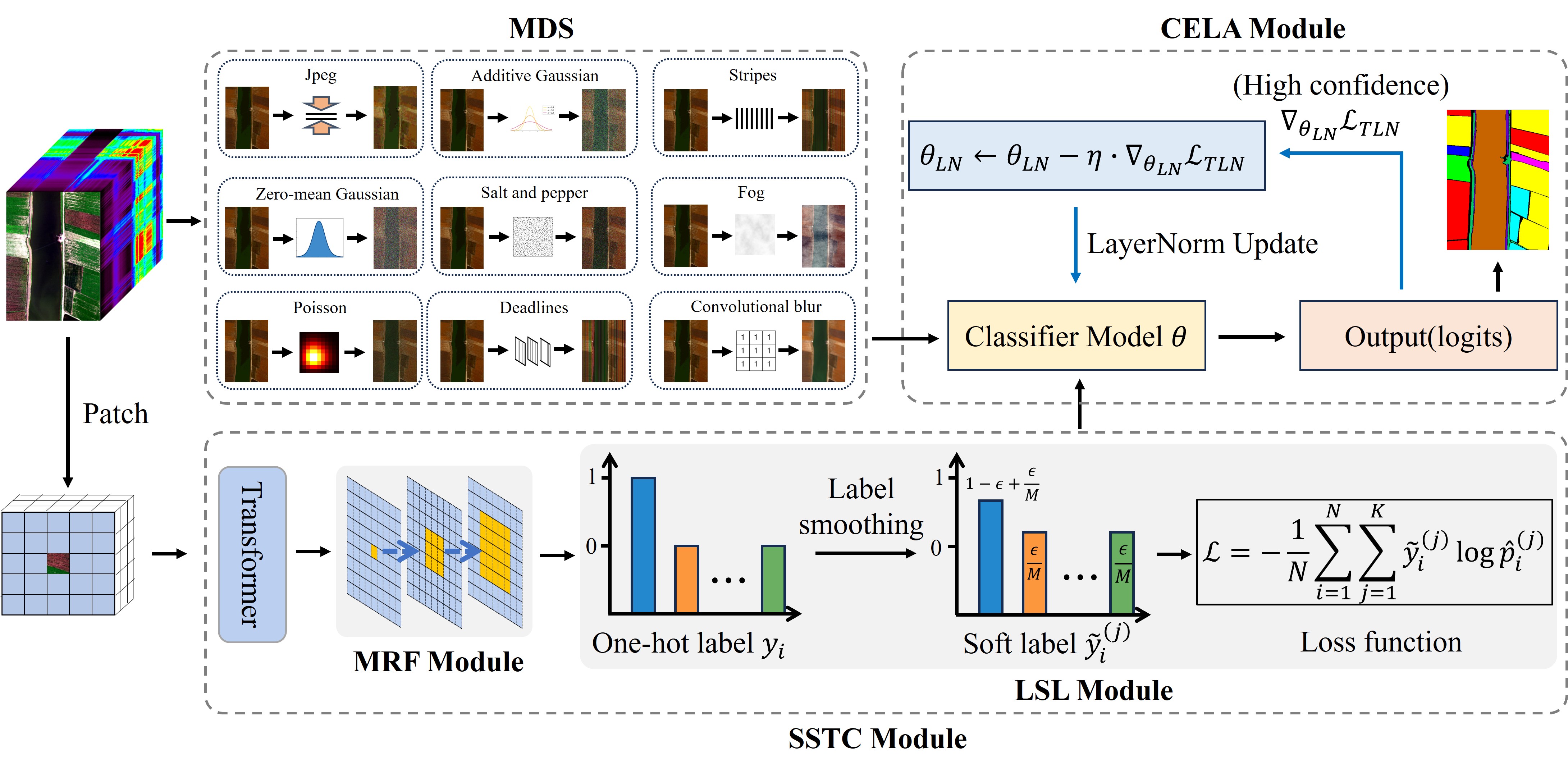}
\caption{Overview of the proposed HyperTTA framework. It consists of three main components: (1) a Multi-Degradation Simulator (MDS)  that constructs unlabeled target-domain inputs by applying nine types of synthetic degradations to clean source data; (2) a Spectral-Spatial Transformer Classifier (SSTC) module trained on clean, labeled samples, incorporating a Multi-level Receptive Field (MRF) module and Label Smoothing Loss (LSL) module for robust representation learning; and (3) a Confidence-aware Entropy-minimized LayerNorm Adapter (CELA) module, which performs TTA by updating only the affine parameters of LayerNorm layers via entropy minimization on high-confidence target samples.}
\label{fig:framework}
\end{figure*}
In this section, we introduce \textbf{HyperTTA} (Test-Time Adaptable Transformer for Hyperspectral Degradation), our proposed framework designed to achieve robust HSI classification under diverse and unknown degradations. As shown in Fig.~\ref{fig:framework}, the overall framework consists of three main components:

\begin{itemize}
    \item \textbf{Multi-Degradation Simulator (MDS):} To simulate realistic distribution shifts, we construct a target-domain dataset using nine representative degradation types (e.g., fog, stripe, Gaussian noise), each controlled by specific parameters. This serves as the unlabeled input domain for TTA.

    \item \textbf{Spectral-Spatial Transformer Classifier (SSTC) Module:} We design a spectral–spatial Transformer architecture that incorporates a multi-level receptive field (MRF) mechanism to capture hierarchical spatial information and applies label smoothing loss (LSL) to improve generalization under noisy conditions. This model is trained only on clean, labeled source data.
    
    \item \textbf{Confidence-aware Entropy-minimized LayerNorm Adapter (CELA) Module:} During inference, we update only the affine parameters (\(\gamma, \beta\)) of Layer Normalization layers by minimizing the prediction entropy on high-confidence unlabeled target samples. This lightweight and selective strategy enables robust and dynamic adaptation to unseen degradations, while avoiding noisy updates, without requiring access to source data or target labels.
\end{itemize}

Together, these components allow HyperTTA to maintain high classification performance across diverse operational scenarios, even in the presence of severe data degradation and domain shifts.

To facilitate the presentation of our method, we first introduce the basic notations and problem setting. Let $\mathcal{X} \in \mathbb{R}^{H \times W \times C}$ denote a HSI, where $H$, $W$, and $C$ represent the height, width, and number of spectral bands, respectively. The corresponding label space is denoted as $\mathcal{Y} = {1, 2, ..., K}$, where $K$ is the total number of semantic classes.

We define the problem under a domain adaptation or TTA setting, where the data distributions differ between the training and testing environments. Let the \textbf{source domain} be defined as
\begin{equation}
\mathcal{D}_s = \left\{ (\mathbf{x}_i^s, y_i^s) \right\}_{i=1}^{N_s},
\end{equation}
where $\mathbf{x}_i^s \in \mathbb{R}^{H \times W \times C}$ denotes the $i$-th clean hyperspectral sample in the source domain, $y_i^s \in \{1, 2, \dots, K\}$ is the corresponding ground-truth label , $N_s$ is the total number of labeled source samples, and $K$ is the number of semantic classes.

The \textbf{target domain} is defined as
\begin{equation}
\mathcal{D}_t = \left\{ \mathbf{x}_j^t \right\}_{j=1}^{N_t},
\end{equation}
where $\mathbf{x}_j^t$ denotes the $j$-th unlabeled sample from the target domain. These samples are typically degraded, and no corresponding labels $y_j^t$ are available during testing or adaptation. $N_t$ denotes the number of target samples.

First, we construct a multi-degradation hyperspectral dataset that systematically simulates and organizes nine representative types of real-world degradations, including noise, blur, compression, and atmospheric effects. This dataset serves as a foundation for evaluating model robustness and supports the development of degradation-aware learning and adaptation techniques. Second, we aim to learn a model $f_\theta$, parameterized by $\theta$, that performs robust classification on the target domain under unknown and potentially degraded conditions. During TTA, the model parameters $\theta$ are updated based solely on the unlabeled target data, with the goal of minimizing prediction uncertainty or aligning feature distributions.

\subsection{Hyperspectral Multi-Degradation Dataset Generation}

\begin{figure}[htbp]
\centering
\includegraphics[width=0.5\textwidth]{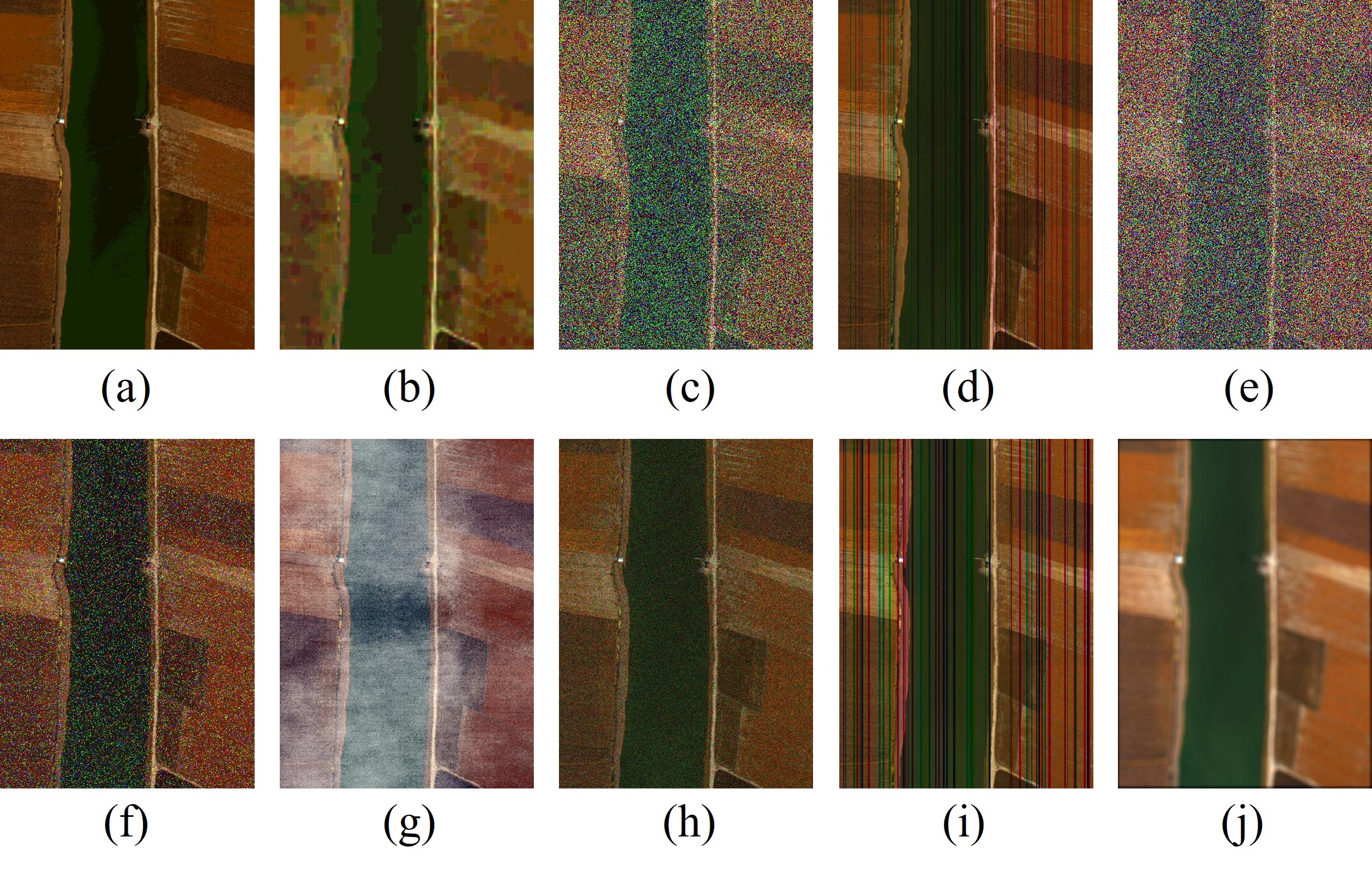}
\caption{False-Color Visualization of a HSI with Multiple Degradations. (a) Clean, (b) Jpeg, (c) Additive Gaussian, (d) Stripes, (e) Zero-mean Gaussian, (f) Salt and pepper, (g) Fog, (h) Poisson, (i) Deadlines, (j) Convolutional Blur.}
\label{fig:multi_degradation}
\end{figure}

To facilitate the study of robust HSI classification under complex and diverse real-world conditions, we construct a multi-degradation hyperspectral dataset that systematically simulates and organizes nine representative types of degradations as shown in Fig.~\ref{fig:multi_degradation}. This dataset serves as a benchmark for evaluating noise-resilient models and TTA techniques.

The degraded sample is represented as $\tilde{\mathcal{X}} = \mathcal{N}_\phi(\mathcal{X})$, where $\mathcal{N}_\phi$ denotes a degradation transformation parameterized by $\phi$. Each transformation simulates a specific degradation source, as detailed below.

\begin{itemize}
    \item \textbf{JPEG Compression.} 
    Simulates compression artifacts caused by data transmission and storage:
    \begin{equation}
    \tilde{\mathcal{X}} = \text{JPEG}_q(\mathcal{X}),
    \end{equation}
    where $q$ is the compression ratio.

    \item \textbf{Zero-Mean Gaussian Noise (ZMGauss).}
    To simulate zero-mean Gaussian noise (ZMGauss), each spectral band is first normalized independently to the $[0, 1]$ range to ensure numerical stability. The normalization for the $c$-th band is computed as:
    \begin{equation}
    \mathcal{X}_{\text{norm}}^{(c)} = \frac{\mathcal{X}^{(c)} - \min(\mathcal{X}^{(c)})}{\max(\mathcal{X}^{(c)}) - \min(\mathcal{X}^{(c)})},
    \label{norm}
    \end{equation}
    where $\mathcal{X}^{(c)} \in \mathbb{R}^{H \times W}$ denotes the $c$-th spectral band of the HSI $\mathcal{X}$. The operation $\min(\mathcal{X}^{(c)})$ and $\max(\mathcal{X}^{(c)})$ denote the minimum and maximum pixel values in that band, respectively. This process rescales each band to the unit interval $[0, 1]$. Subsequently, zero-mean Gaussian noise with fixed variance is added to each normalized band:
    \begin{equation}
    \tilde{\mathcal{X}}^{(c)} = \text{clip} \left( \mathcal{X}_{\text{norm}}^{(c)} + \epsilon \right), \quad \epsilon \sim \mathcal{N}(0, \sigma^2),
    \end{equation}
    where $\epsilon$ is a noise matrix of the same shape as $\mathcal{X}^{(c)}$, sampled independently from a Gaussian distribution with zero mean and variance $\sigma^2$. The function $\text{clip}(\cdot)$ ensures that all pixel values remain within the $[0, 1]$ range after noise addition.

    \item \textbf{Additive Gaussian Noise.}
    To simulate band-specific additive Gaussian noise, we first normalize each spectral band of the clean HSI to the $[0, 1]$ range as (\ref{norm}). Then, for each band, we sample a standard deviation $\sigma_c$ from a uniform distribution:
    \begin{equation}
    \sigma_c \sim \mathcal{U}(0, \sigma_{\text{max}}),
    \end{equation}
    where $\sigma_{\text{max}}$ controls the maximum allowed noise intensity. Next, we generate a noise matrix $\epsilon_c \in \mathbb{R}^{H \times W}$ whose entries are drawn i.i.d. from a Gaussian distribution with zero mean and variance $\sigma_c^2$:
    \begin{equation}
    \epsilon_c \sim \mathcal{N}(0, \sigma_c^2).
    \end{equation}
    Then, the noisy band is obtained by:
    \begin{equation}
    \tilde{\mathcal{X}}^{(c)} = \text{clip} \left( \mathcal{X}_{\text{norm}}^{(c)} + \epsilon_c \right),
    \end{equation}
    where $\text{clip}(\cdot)$ ensures that pixel values remain within the $[0, 1]$ range. This process introduces a different noise level per band, thereby better reflecting the non-uniform sensor noise patterns commonly observed in real-world hyperspectral systems.

    \item \textbf{Poisson Noise.} 
    To simulate photon-counting noise that commonly arises in low-light or short-exposure imaging conditions, we apply band-wise Poisson noise to the clean HSI. Since the Poisson distribution requires input values to represent photon counts (non-negative integers), we scale each band appropriately based on a desired signal-to-noise ratio (SNR). First, the $c$-th spectral band $\mathcal{X}^{(c)} \in \mathbb{R}^{H \times W}$ is scaled by a factor $\gamma_c$:
    \begin{equation}
    \mathcal{X}_{\text{scaled}}^{(c)} = \gamma_c \cdot \mathcal{X}^{(c)},
    \end{equation}
    where the scaling factor $\gamma_c$ is computed to match a target SNR level in decibels:
    \begin{equation}
    \gamma_c = \frac{\text{SNR}_{\text{lin}}}{\frac{1}{N} \sum_{i=1}^N \frac{(\mathcal{X}_i^{(c)})^2}{\mathcal{X}_i^{(c)} + \varepsilon}},
    \end{equation}
    with $\text{SNR}_{\text{lin}} = 10^{\text{SNR}_{\text{dB}} / 10}$ being the linear form of the target SNR (in dB), and $N = H \times W$ is the number of pixels in the image. The term $\varepsilon > 0$ is a small constant added to avoid division by zero. Once scaled, each pixel value is used as the mean of a Poisson distribution to generate noisy observations:
    \begin{equation}
    \tilde{\mathcal{X}}^{(c)} \sim \text{Poisson}(\mathcal{X}_{\text{scaled}}^{(c)}),
    \end{equation}
    where the Poisson noise is applied independently to each pixel in the $c$-th band.

    \item \textbf{Salt-and-Pepper Noise.} 
    This type of impulse noise simulates pixel corruption caused by random bit errors or faulty sensors. It randomly replaces a portion of pixels with minimum or maximum values (i.e., ``pepper'' and ``salt'' noise, respectively). Each pixel $(i,j)$ in the $c$-th normalized spectral band is corrupted independently with probability $p$, and remains unchanged with probability $1 - p$. The noisy value is defined as:
    \begin{equation}
    \tilde{\mathcal{X}}_{i,j}^{(c)} =
    \begin{cases}
    0, & \text{with probability } \frac{p}{2}, \\
    1, & \text{with probability } \frac{p}{2}, \\
    (\mathcal{X}_{i,j}^{(c)})_{\text{norm}}, & \text{with probability } 1 - p,
    \end{cases}
    \end{equation}
    where $(\mathcal{X}_{i,j}^{(c)})_{\text{norm}}$ is the normalized clean pixel value in the $c$-th band, and $p \in [0,1]$ controls the noise ratio. This model ensures that approximately $p \cdot 100\%$ of the pixels in each band are corrupted: half are replaced with the minimum value (0, ``pepper''), and the other half with the maximum value (1, ``salt''). The corruption is applied independently to each spectral band to simulate spectrum-wise random dropout.

   \item \textbf{Stripe Noise.}
    To simulate systematic stripe-like degradation commonly observed in hyperspectral sensors (e.g., pushbroom scanners), we inject vertical stripe noise at random column positions in each spectral band. For each spectral band $c \in \{1, \dots, C\}$, we first randomly sample the number of stripe columns:
    \begin{equation}
    n_c \sim \mathcal{U}_{\text{int}}(\text{a}, \text{b}),
    \label{range}
    \end{equation}
    where $\mathcal{U}_{\text{int}}(a,b)$ denotes a discrete uniform distribution over integers in $[a, b)$. Then, a set of stripe column indices $\mathcal{J}_c \subset \{1, \dots, W\}$ with $|\mathcal{J}_c| = n_c$ is sampled uniformly without replacement.  For each $j \in \mathcal{J}$, all pixels in the $j$-th column are overwritten with a fixed intensity value $\lambda$, independently for each band:
    \begin{equation}
    \tilde{\mathcal{X}}^{(c)}_{:, j} = \lambda_c, \quad \lambda_c \sim \mathcal{U}(0.6, 0.8),
    \end{equation}
    where $\lambda_c$ is sampled independently for each band $c$ to simulate spectrally inconsistent stripe patterns. This process is repeated across all spectral bands, resulting in a non-uniform stripe pattern across the entire cube. Note that the stripe intensity $\lambda_c$ is randomly chosen from a fixed range $[0.6, 0.8]$ after band normalization.

    \item \textbf{Deadline Noise.}
    Deadline noise simulates sensor defects in hyperspectral imaging systems, where some sensor rows or columns permanently fail and produce invalid or zero-valued readings. To avoid accessing columns beyond the image boundary, we restrict the starting positions of the deadlines to $\mathcal{D}_c \subset \{1, \dots, W - 3\}$, where each $j \in \mathcal{D}_c$ denotes the starting column of a dead stripe, assuming that the maximum stripe width is 3. The number of such positions is randomly determined as (\ref{range}) and $|\mathcal{D}_c| = n_c$. For each $j \in \mathcal{D}_c$, we randomly assign a stripe width $w_j \in \{1, 2, 3\}$, representing the number of consecutive columns set to zero. The final degradation is applied as:
    \begin{equation}
    \tilde{\mathcal{X}}^{(c)}_{:, j : j + w_j - 1} = 0, \quad \forall j \in \mathcal{D}_c.
    \end{equation}
    Here, the slice notation $j : j + w_j - 1$ denotes the set of consecutive columns to be zeroed out, with width $w_j$. The colon operator $:$ refers to the full spatial height, meaning that the operation sets all pixels in the selected columns to zero, across the entire image height.
    \par This operation replaces all pixel values in the selected vertical stripes with zero, creating abrupt blacked-out columns that persist across the full height of the image. Unlike stripe noise, which replaces column values with moderate random intensities to simulate illumination variation or sensor drift, deadline noise sets entire columns to zero with varying width to emulate complete sensor failure. The effect is more severe and disruptive, resembling hardware-level defects rather than soft degradations.

    \item \textbf{Convolutional Blur.}
    To simulate defocus or motion blur that may occur during imaging, we apply a spatial convolution to each spectral band using a uniform mean filter. For each band $c \in \{1, \dots, C\}$, the blurred image is generated as:
    \begin{equation}
    \tilde{\mathcal{X}}^{(c)} = \mathcal{X}^{(c)} * K,
    \end{equation}
    where $*$ denotes the 2D convolution operation and $K \in \mathbb{R}^{k \times k}$ is a normalized mean kernel defined as:
    \begin{equation}
    K = \frac{1}{k^2} \mathbf{1}_{k \times k} ,
    \end{equation}
    where $\mathbf{1}_{k \times k}$ represents a $k \times k$ matrix filled with ones. The convolution kernel $K$ performs uniform averaging over a local spatial window, which effectively smooths high-frequency details and edges in each band. This process mimics the loss of sharpness caused by optical blur, camera shake, or lens defocus. The kernel size $k$ determines the strength of the blur: a larger $k$ leads to a stronger smoothing effect. The operation is independently applied to each spectral band without cross-band mixing, preserving spectral locality while degrading spatial detail.

    \item \textbf{Fog Degradation.}
    To simulate atmospheric scattering effects in hazy scenes, we follow a wavelength-aware optical model that synthesizes fog on HSI inspired by \cite{9134800}. For each spectral band $c$, the degraded image is generated as:
    \begin{equation}
    \tilde{\mathcal{X}}^{(c)} = \mathcal{X}^{(c)} \cdot t^{(c)} + A^{(c)} \cdot (1 - t^{(c)}),
    \end{equation}
    where $\mathcal{X}^{(c)}$ is the clean input at band $c$, $A^{(c)}$ is the estimated global atmospheric light, and $t^{(c)}$ is the wavelength- and location-dependent transmission map. The transmission is modeled as:
    \begin{equation}
    t^{(c)}(x,y) = \left( t_1(x,y) \right)^{\left( \frac{\lambda_0}{\lambda_c} \right)^{\gamma(x,y)}},
    \end{equation}
    with $t_1(x,y) = 1 - \omega \cdot \rho(x,y)$ being the base transmittance map derived from normalized fog density, and $\gamma(x,y)$ being an empirically regressed decay factor fitted over reference reflectance-transmittance pairs. Here, $\lambda_c$ is the center wavelength of band $c$, and $\lambda_0$ is the shortest wavelength used as reference. This formulation introduces both spatial variation (via $t_1$ and $\gamma$) and spectral decay (via $\lambda_c$) to realistically model wavelength-dependent visibility in foggy conditions.
\end{itemize}

All degradations are applied independently across multiple benchmark HSI datasets (e.g., Pavia, WHU-Longkou), resulting in a comprehensive multi-degradation dataset for robust model development and evaluation. For each dataset, we generate and store:
\begin{itemize}
    \item Degraded HSIs for each noise type;
    \item Parameter metadata for reproducibility;
    \item False-color visualizations for qualitative assessment.
\end{itemize}

For all degradation types, we first normalize each spectral band to the range $[0,1]$ before applying noise. The resulting degraded images may then be rescaled back to the original dynamic range of the clean images if required by downstream tasks such as classification. This dataset provides a standardized platform for benchmarking models under various degradation scenarios, and supports future research on noise-aware learning, domain adaptation, and test-time robustness in HSI analysis.

\subsection{Spectral-Spatial Transformer Framework with Multi-Level Receptive Field and Label Smoothing}

We propose a spectral–spatial classification framework that avoids PCA-based band compression. Instead, it incorporates a multi-level receptive field mechanism and label smoothing regularization to improve generalization. The multi-level receptive field enables the model to capture both fine-grained local textures and broader contextual dependencies, which is crucial for handling diverse degradation patterns. Meanwhile, label smoothing mitigates overconfidence in predictions and improves model robustness under noisy or ambiguous inputs, thereby enhancing adaptation to multi-degradation scenarios.

\subsubsection{Input and Patch Construction}
For each central pixel, we extract a local patch \(\mathbf{x}_{i}^{s} \in \mathbb{R}^{C \times w \times w}\), where \(w\) is the patch size. Each sample is associated with a label \(y_i \in \{1, 2, \dots, K\}\), and the training set contains \(N\) such pairs.

\subsubsection{Multi-Level Receptive Field Mechanism}
To capture spatial context under different scales, we employ a multi-branch structure to extract features from the patch \(\mathbf{x}_{i}^{s}\) using convolutional kernels of multiple sizes. Specifically, for each branch \(m = 1, \dots, M\), we compute:
\begin{equation}
\mathbf{F}_i^{(m)} = \text{Conv}_{k_m}(\mathbf{x}_{i}^{s}),
\end{equation}

where \(k_m\) is the convolution kernel size (e.g., \(k_1=3\), \(k_2=5\), \(k_3=7\)) and \(\mathbf{F}_i^{(m)} \in \mathbb{R}^{d_m \times w' \times w'}\) is the resulting feature map. Each output is projected to a fixed dimension:
\begin{equation}
\mathbf{F}_i = \text{Concat}\left( \phi_1(\mathbf{F}_i^{(1)}), \dots, \phi_M(\mathbf{F}_i^{(M)}) \right) \in \mathbb{R}^{d \times w' \times w'},
\end{equation}
where \(\phi_m(\cdot)\) denotes a \(1 \times 1\) convolution for dimension alignment, and \(d = \sum_{m=1}^M d_m'\) is the total fused feature dimension. The combined feature map is flattened into a token sequence:
\begin{equation}
\mathbf{Z}_i = \text{Flatten}(\mathbf{F}_i) \in \mathbb{R}^{L \times d}, \quad L = w'^2.
\end{equation}

\subsubsection{Transformer Encoding}
The token sequence \(\mathbf{Z}_i\) is passed through a stack of Transformer encoder layers with multi-head self-attention. For each attention block, we compute:
\begin{equation}
\text{Attention}(\mathbf{Q}, \mathbf{K}, \mathbf{V}) = \text{softmax}\left( \frac{\mathbf{QK}^\top}{\sqrt{d}} \right) \mathbf{V},
\end{equation}
where \(\mathbf{Q}, \mathbf{K}, \mathbf{V} \in \mathbb{R}^{L \times d}\) are linearly projected query, key, and value matrices. The final token corresponding to the center position is extracted as the classification token \(\mathbf{z}_i^{\text{final}} \in \mathbb{R}^{d}\) and fed into the classifier head:
\begin{equation}
\hat{\mathbf{p}}_i = \text{softmax}( \mathbf{W}_{\text{cls}} \mathbf{z}_i^{\text{final}} + \mathbf{b}_{\text{cls}} ),
\end{equation}
where \(\hat{\mathbf{p}}_i \in \mathbb{R}^{K}\) is the predicted class distribution and \(\mathbf{W}_{\text{cls}}, \mathbf{b}_{\text{cls}}\) are trainable parameters.

\subsubsection{Label Smoothing Loss}
To avoid overconfident predictions and improve model generalization, we adopt label smoothing regularization~\cite{szegedy2016rethinking}. For each label \(y_i\), we define a smoothed target distribution \(\tilde{\mathbf{y}}_i \in \mathbb{R}^K\) as:
\begin{equation}
\tilde{y}_i^{(j)} =
\begin{cases}
1 - \epsilon + \frac{\epsilon}{K}, & \text{if } j = y_i, \\
\frac{\epsilon}{K}, & \text{otherwise},
\end{cases}
\end{equation}
where \(\epsilon \in [0, 1]\) is the smoothing factor, and \(\tilde{y}_i^{(j)}\) is the smoothed label assigned to class \(j\). The total training loss is computed as:
\begin{equation}
\mathcal{L} = -\frac{1}{N} \sum_{i=1}^{N} \sum_{j=1}^{K} \tilde{y}_i^{(j)} \log \hat{p}_i^{(j)}.
\end{equation}

\subsection{TTA with Entropy Minimization and Confidence Filtering}

To improve the robustness of the model under distribution shift without access to source data, we adopt a TTA strategy based on entropy minimization~\cite{wang2020tent}, enhanced by a confidence-based sample selection mechanism. Specifically, we update only the affine parameters—namely, the scale \(\gamma\) and shift \(\beta\)—of the Layer Normalization (LayerNorm) layers in the Transformer, while keeping all other model parameters frozen. This enables the model to dynamically adapt to target-domain statistics during inference, without requiring additional training or target labels.

Let \(\mathbf{x}_j^t\) denote an unlabeled target-domain input sample, and let \(f_\theta\) denote the pre-trained classification model parameterized by \(\theta\). During test time, we seek to update a subset of parameters \(\theta_{\text{LN}} \subset \theta\) corresponding to \(\{\gamma, \beta\}\) in each LayerNorm layer, such that the model minimizes prediction uncertainty. To quantify uncertainty, we use the entropy of the softmax output:
\begin{equation}
\mathcal{H}(f_\theta(\mathbf{x}_j^t)) = - \sum_{k=1}^{K} \hat{p}_j^{(k)} \log \hat{p}_j^{(k)},
\end{equation}
where \(\hat{p}_j^{(k)}\) is the predicted probability for class \(k\) given the input \(\mathbf{x}_j^t\).

To ensure stable and effective adaptation, we employ a hybrid confidence selection strategy. Let the confidence of sample \(\mathbf{x}_j^t\) be defined as:
\begin{equation}
c_j = \max_{k} \hat{p}_j^{(k)},
\end{equation}
and define a confidence threshold \(\tau \in [0, 1]\). Let \(\mathcal{I}_\tau = \{ j \mid c_j > \tau \}\) be the set of samples above the threshold. To avoid the case where too few samples are selected (i.e., \(|\mathcal{I}_\tau|\) is small), we also define \(\mathcal{I}_{\text{top}} \subset \{1, \dots, B\}\) as the indices corresponding to the top 30\% most confident samples in the mini-batch. We then define the final selected index set as:
\begin{equation}
\mathcal{I} = 
\begin{cases}
\mathcal{I}_\tau, & \text{if } |\mathcal{I}_\tau| \geq 0.3B \\
\mathcal{I}_{\text{top}}, & \text{otherwise}
\end{cases}
\end{equation}

The adaptation loss is computed only on the selected samples:
\begin{equation}
\mathcal{L}_{\text{LN}} = \frac{1}{|\mathcal{I}|} \sum_{j \in \mathcal{I}} \mathcal{H}(f_\theta(\mathbf{x}_j^t)).
\end{equation}

During inference, for each mini-batch \(\{\mathbf{x}_j^t\}_{j=1}^{B}\), we perform \(S\) steps of gradient descent to update only \(\gamma\) and \(\beta\), while freezing all other parameters. Specifically, the updates are:
\begin{equation}
\gamma \leftarrow \gamma - \eta \cdot \frac{\partial \mathcal{L}_{\text{LN}}}{\partial \gamma}, \quad
\beta \leftarrow \beta - \eta \cdot \frac{\partial \mathcal{L}_{\text{LN}}}{\partial \beta},
\end{equation}
where \(\eta\) is the learning rate for test-time optimization.

We adopt an episodic adaptation strategy: for each test sample or batch, the LayerNorm parameters \(\{\gamma, \beta\}\) are reset to their original pre-trained values before adaptation, ensuring that each test instance is adapted independently and preventing parameter drift. Let \(\theta_0\) be the original model parameters. We construct an adapted model \(\tilde{f}_\theta\) with the following adaptation scheme:
\begin{itemize}
    \item All parameters are initialized to \(\theta_0\);
    \item Gradients are enabled only for \(\gamma\) and \(\beta\) in LayerNorm layers;
    \item For each mini-batch, only high-confidence samples (based on threshold \(\tau\) or top-30\%) are used to compute the entropy loss for adaptation.
\end{itemize}

\section{Experiments}
\subsection{Experimental Settings}

\subsubsection{Datasets}

\begin{figure}[htbp]
\centering
\includegraphics[width=0.5\textwidth]{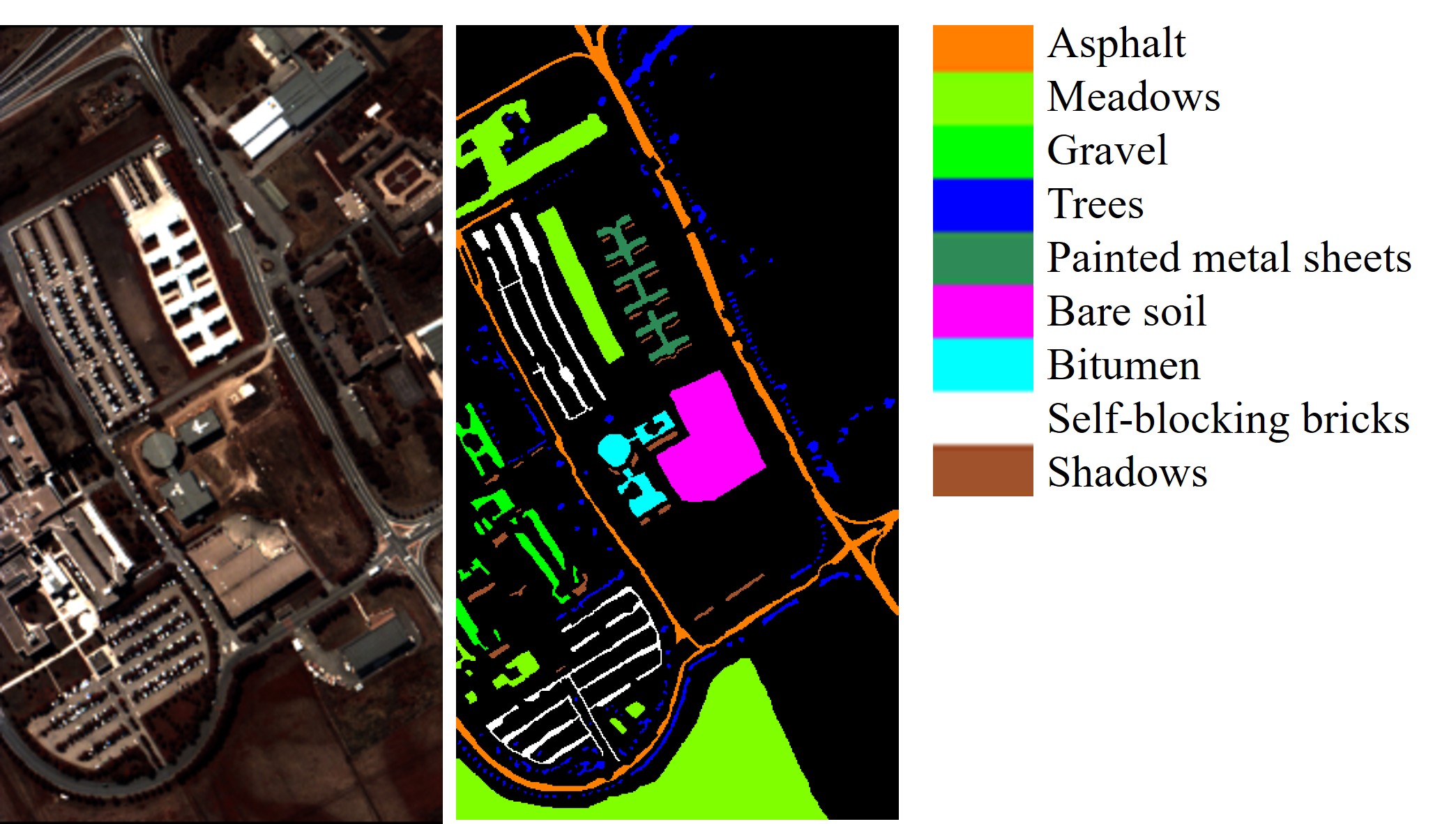}
\caption{Visualization of the Pavia University dataset: false-color composite image, ground-truth map, and class legend.}
\label{fig:pavia}
\end{figure}

\begin{figure}[htbp]
\centering
\includegraphics[width=0.5\textwidth]{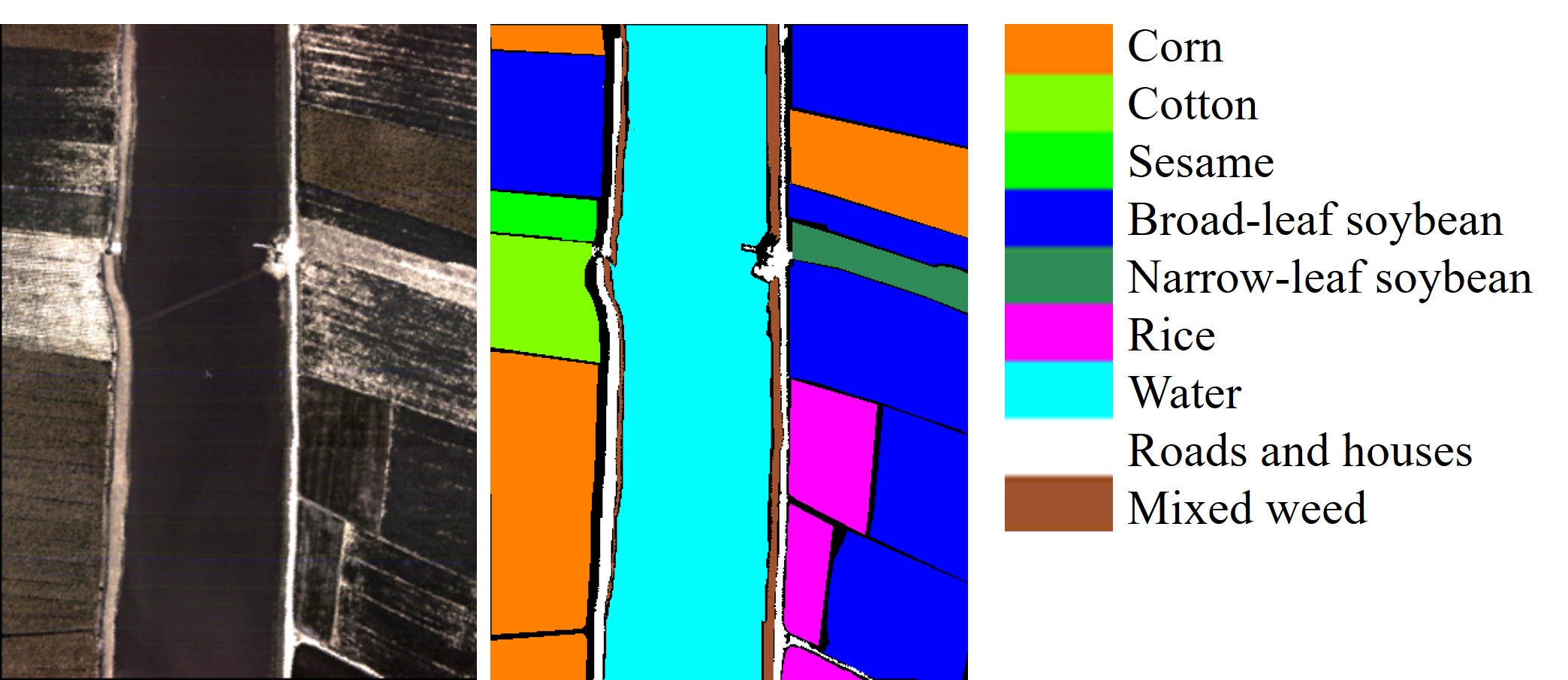}
\caption{Visualization of the WHU-Hi-Longkou dataset: false-color composite image, ground-truth map, and class legend.}
\label{fig:longkou}
\end{figure}

To verify the effectiveness of our algorithm,we applied the algorithm to two public datasets: the Pavia University dataset (PU) and the WHU-Hi-LongKou dataset (WHLK):
\begin{itemize}
     \item The Pavia University dataset as shown in Fig.~\ref{fig:pavia}, captured by the Reflective Optics System Imaging Spectrometer (ROSIS) sensor in 2003, covers an urban area surrounding the University of Pavia, Italy. This dataset comprises 115 spectral bands spanning wavelengths from 430 to 860 nm, with a spatial resolution of 1.3 meters and a spatial dimension of 610 × 340 pixels. It includes 9 distinct land-cover classes such as asphalt, meadows, trees, and metal sheets. For experimental purposes, 12 noise-affected bands were typically removed, resulting in 103 usable spectral bands. The dataset has been widely adopted for urban land-use classification studies due to its detailed representation of man-made structures and vegetation.
    \item The WHU-Hi-LongKou dataset~\cite{ZHONG2020112012,8573977} as shown in Fig.~\ref{fig:longkou}, was obtained 13:49 to 14:37 on July 17, 2018, in Longkou Town, Hubei province, China.This dataset was captured using a Headwall Nano-Hyperspec imaging sensor with a 8-mm focal length, mounted on a DJI Matrice 600 Pro unmanned aerial vehicle (UAV) operating at an altitude of 500 m. The resulting images,with a resolution of 550 × 400 pixels,encompass 270 spectral bands ranging from 400 to 1000 nm, offering a spatial resolution of approximately 0.463 m for the hyperspectral UAV imagery. The captured scene represents a complex agricultural landscape, featuring various crop types, such as Corn, Cotton and Sesame. In total, the dataset covers 9 distinct land-cover classes. 
\end{itemize}
The detailed specifications of the two datasets are provided in Table~\ref{tab:detail}.

\begin{table}[htbp]
\small
\begin{center}
    \caption{Detailed data of the PU dataset and the WHLK dataset}
    \label{tab:detail}
    \begin{adjustbox}{width=0.5\textwidth}
\begin{tabular}{c|lr|lr}
\toprule
\multirow{2}{*}{Class} & \multicolumn{2}{c|}{\textbf{Pavia University}}  & \multicolumn{2}{c}{\textbf{WHU-Hi-LongKou}} \\
& Class Name & Samples & Class Name & Samples \\
\midrule
1 & Asphalt & 6631 &  Corn & 34511 \\
2 & Meadows & 18649 &  Cotton & 8374 \\
3 & Gravel & 2099 &  Sesame & 3031 \\
4 & Trees & 3064 & Broad-leaf soybean & 63212 \\
5 & PaintedMetalSheets & 1345  & Narrow-leaf soybean & 4151 \\
6 & BareSoil & 5029 &  Rice & 11854 \\
7 & Bitumen & 1330 & Water & 67056 \\
8 & SelfBlockingBricks & 3682  & Roads and houses & 7124 \\
9 & Shadows & 947 &  Mixed weed & 5229 \\
\midrule
Total & & 42776 & & 204542\\
\bottomrule 
\end{tabular} 
\end{adjustbox}
\end{center}
\end{table}

\subsubsection{Comparison Method}

To evaluate the effectiveness of the proposed method, we compare it with several representative baselines and state-of-the-art TTA techniques. SSRN~\cite{8061020} (Spectral–Spatial Residual Network) serves as a strong fully supervised baseline without any adaptation. SSFTT~\cite{9684381} is a Transformer-based model that incorporates spectral–spatial fusion and is trained under a fully supervised setting, serving as an advanced backbone for comparison. CASST~\cite{9874815} is a Transformer-based architecture that incorporates spatial and channel attention mechanisms to enhance local feature modeling for HSI classification. MASSFormer~\cite{10506482} employs multi-scale attention to jointly model spectral and spatial features, achieving robust performance under various degradations.Tent~\cite{wang2020tent} adapts the model during inference by minimizing the entropy of predictions on unlabeled target samples, without access to source data. SoTTA~\cite{gong2023sotta} improves robustness by introducing confidence-based filtering and entropy-sharpness minimization, and adapts only the parameters of LayerNorm during testing. CoTTA~\cite{wang2022continual} performs test-time self-training with EMA modeling and stochastic restoration to enable continual adaptation under distribution shifts. SAR~\cite{niu2023towards} uses structure-aware reconstruction to preserve spatial-spectral consistency for robust domain adaptation in HSI classification. CFA~\cite{kojima2022robustvit} enhances the robustness of Vision Transformers at test time by introducing a consistency-based filtering approach. BNAdapt~\cite{LI2018109} is a simple TTA method that updates Batch Normalization statistics using target data, enabling domain adaptation without requiring source labels.

\subsubsection{Implementation Details}

All experiments were conducted on a workstation equipped with a 13th Gen Intel(R) Core(TM) i7-13700KF CPU (3.4 GHz, 16 cores), 64 GB of RAM, and an NVIDIA GeForce RTX 3090 GPU with 24 GB of memory. The proposed model was implemented using PyTorch. During the initial training phase, the patch and batch sizes were set to 15 and 64 for the PU dataset, and 21 and 32 for the WHLK dataset, respectively. The training was conducted using the Adam optimizer with a learning rate of 0.001, for 20 epochs on the PU dataset and 10 epochs on the WHLK dataset. And the smoothing factor \(\epsilon=0.05\). For supervised training, 20\% of the clean samples and their corresponding labels from each class were used. The remaining 80\% were transformed into degraded inputs using our proposed multi-degradation simulation strategy, and their labels were discarded during adaptation. These degraded samples were then used as TTA inputs. During the TTA phase, model parameters were updated for a single epoch using the stochastic gradient descent (SGD) optimizer with a learning rate of 0.001 and a batch size of 64. To ensure the reliability of the adaptation process, only high-confidence target samples—selected based on a confidence threshold of \(\tau=0.8\)—were utilized for parameter updates. All experimental results are reported as the average over 3 independent runs to ensure statistical robustness. The specific degradation simulation settings for both the PU and WHLK datasets are summarized in Table~\ref{tab:degradation_parameters}.

\begin{table*}
\centering
\caption{Parameters used for simulating nine types of degradation on the PU and WHLK datasets.}
\label{tab:degradation_parameters}
\begin{tabular}{llp{6.5cm}cc}
\toprule
        Degradation Type &  Parameter &                                        Meaning & PU Value & WHLK Value \\
\midrule
        JPEG Compression &                 \(q\) & Compression ratio &                                        110 &         15 \\
 Additive Gaussian Noise &          $\sigma$ &           Standard deviation of additive noise &           0.2 &        0.25 \\
   Salt-and-Pepper Noise &                 \(p\) &      Noise density (probability of corruption) &           0.1 &       0.09 \\
      Convolutional Blur &                 \(k\) &                       Kernel size for blurring &             3 &          3 \\
Zero-Mean Gaussian Noise &          $\sigma$ &          Standard deviation of zero-mean Gaussian noise &          0.25 &       0.15 \\
           Poisson Noise &               SNR &                  Signal-to-noise ratio (in dB) &           -10 &         15 \\
            Stripe Noise &               \(a,b\) &                         Range of stripe counts &         30,35 &      35,40 \\
          Deadline Noise &               \(a,b\) &                         Range of dead columns  &         30,35 &      25,30 \\
         Fog Degradation &          $\omega$ &             Atmospheric scattering coefficient &           0.3 &        0.2 \\
\bottomrule
\end{tabular}
\end{table*}

\begin{figure}[htbp]
\centering
\includegraphics[width=0.5\textwidth]{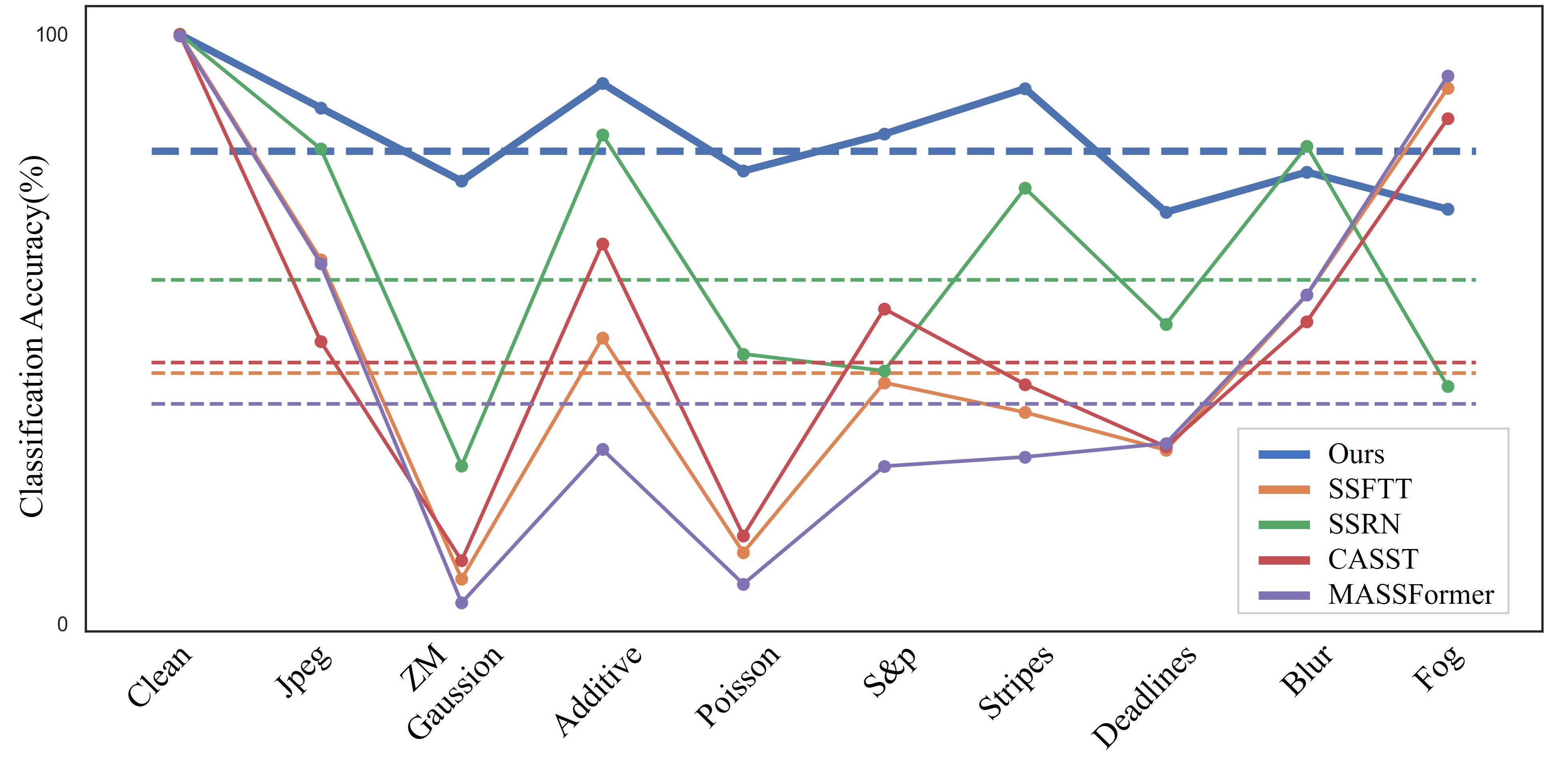}
\caption{Classification accuracy (OA) of different models under various degradation types on the PU dataset. 
Solid lines denote per-degradation performance, while dashed lines represent the average OA across all degradations.}
\label{fig:pu_visual1}
\end{figure}

\begin{figure}[htbp]
\centering
\includegraphics[width=0.5\textwidth]{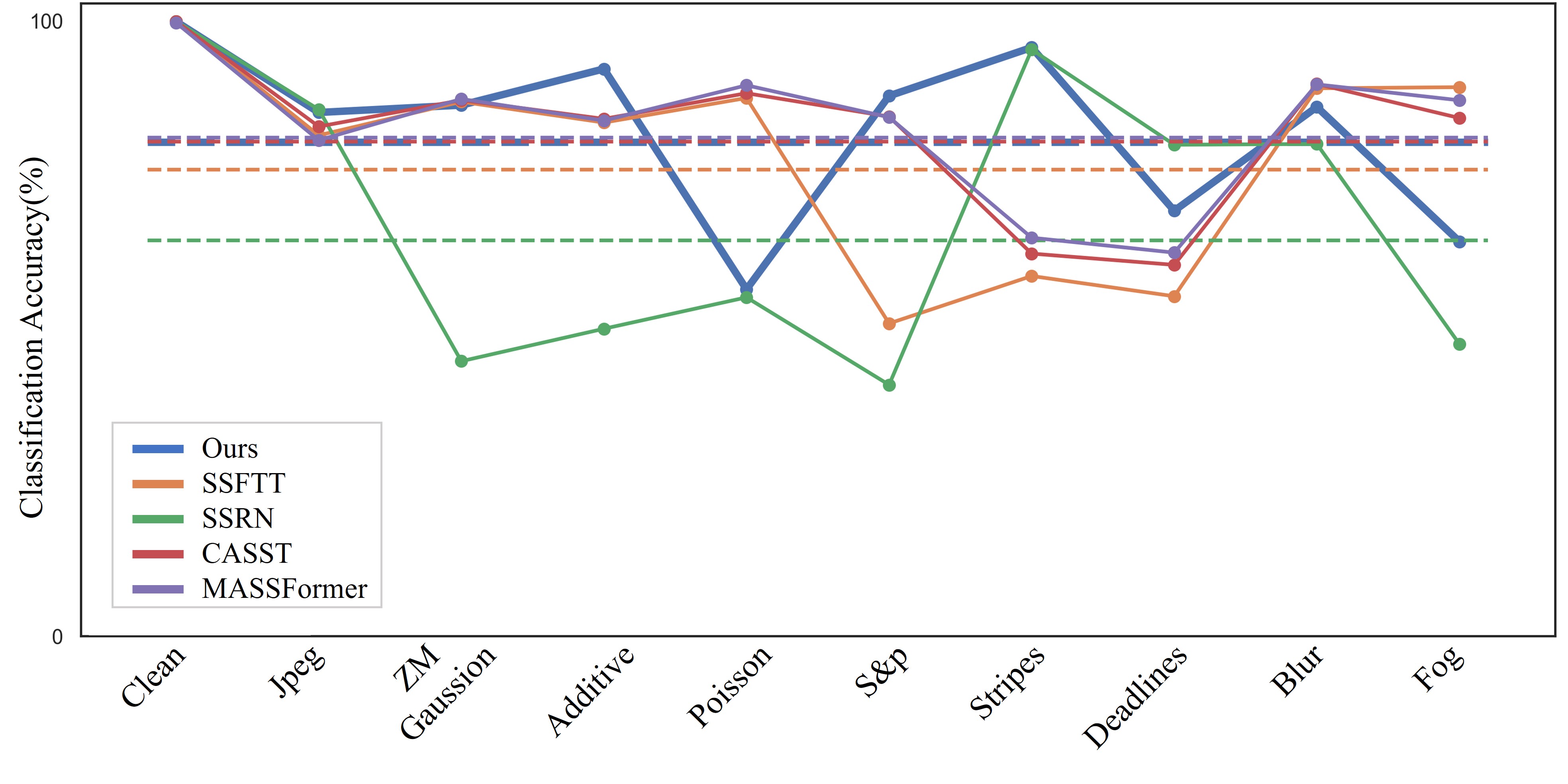}
\caption{Classification accuracy (OA) of different models under various degradation types on the WHLK dataset. 
Solid lines denote per-degradation performance, while dashed lines represent the average OA across all degradations.}
\label{fig:whlk_visual1}
\end{figure}

\begin{figure}[htbp]
\centering
\includegraphics[width=0.5\textwidth]{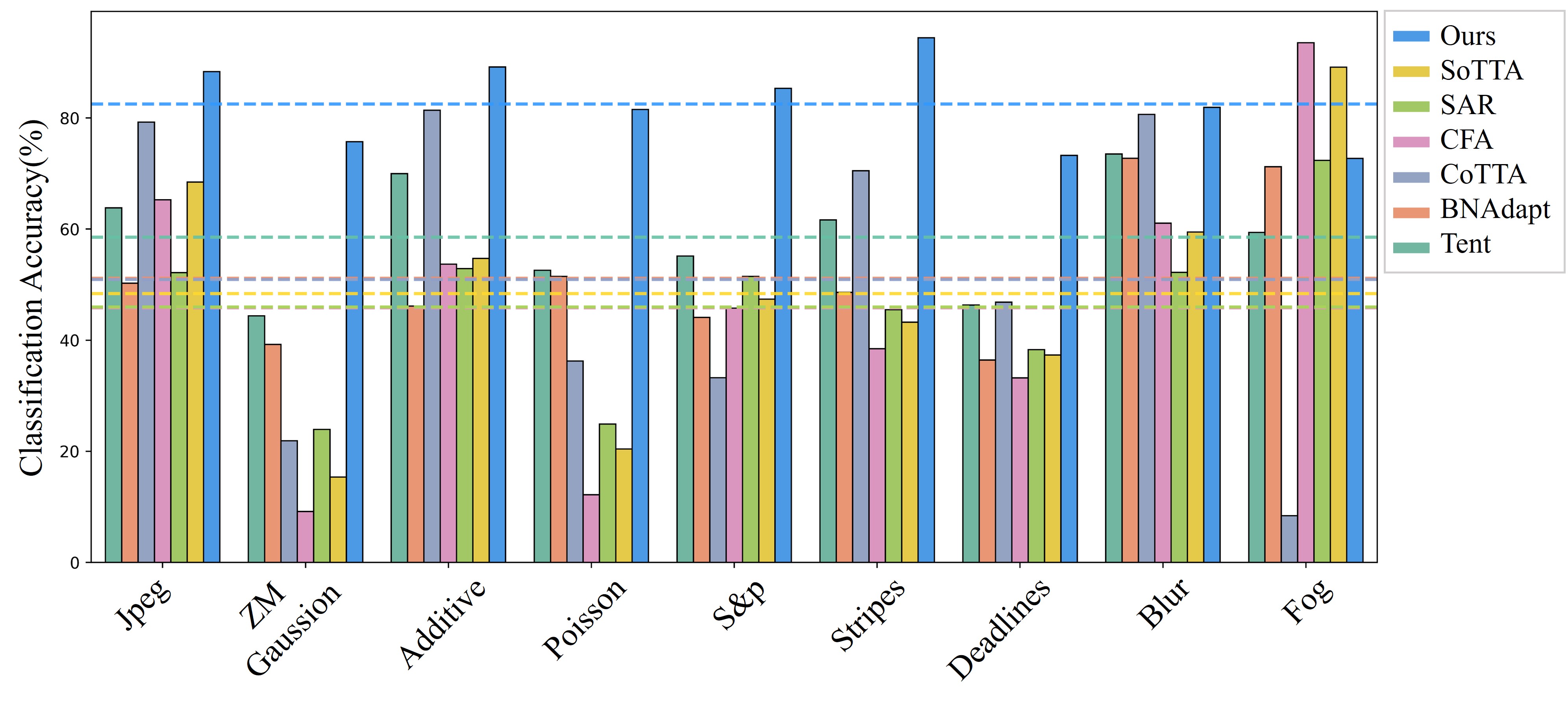}
\caption{Classification performance (OA) of different TTA methods under diverse degradation types on the PU dataset. Each group of bars corresponds to a specific degradation type, and the dashed lines represent the average accuracy of each method across all degradations.}
\label{fig:PU_visual3}
\end{figure}

\begin{figure}[htbp]
\centering
\includegraphics[width=0.5\textwidth]{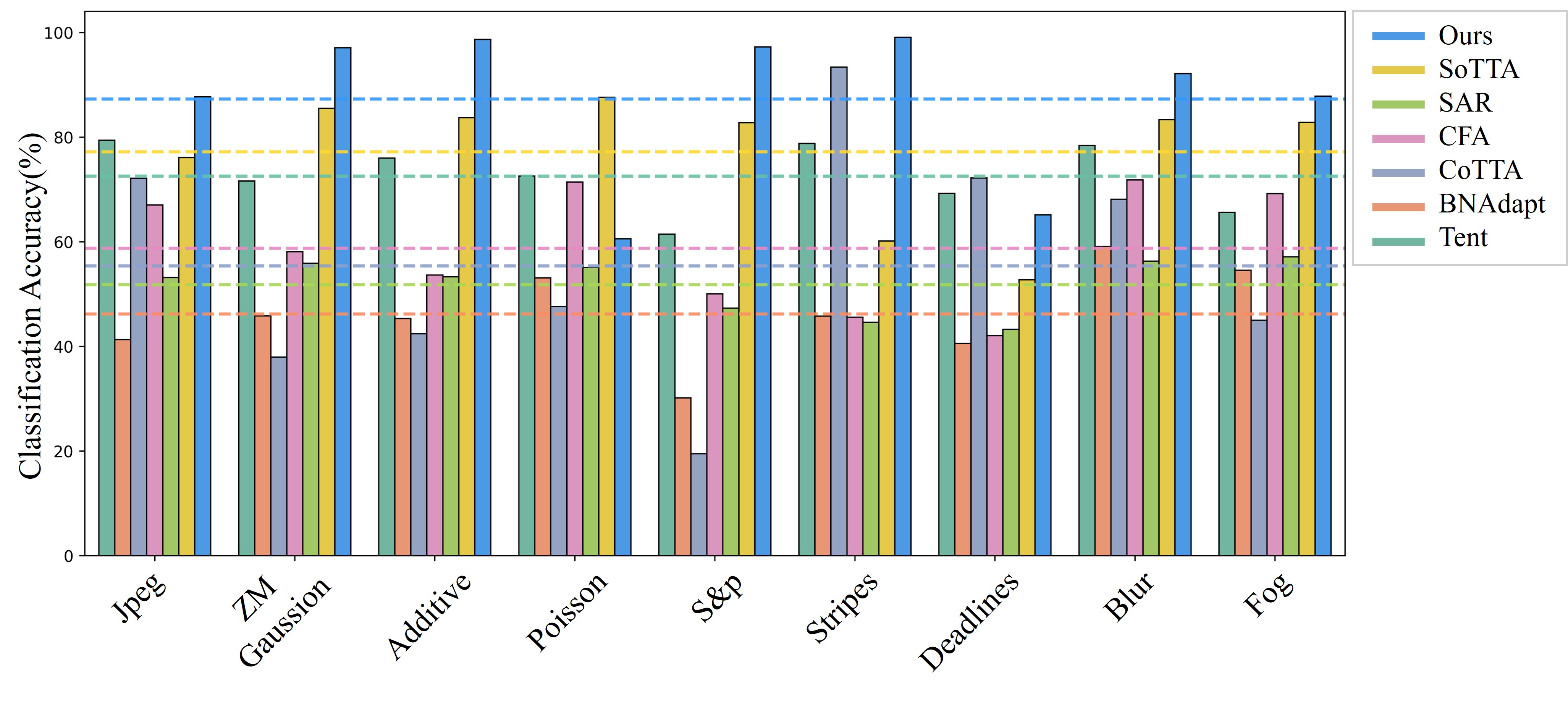}
\caption{Classification performance (OA) of different TTA methods under diverse degradation types on the WHLK dataset. Each group of bars corresponds to a specific degradation type, and the dashed lines represent the average accuracy of each method across all degradations.}
\label{fig:whlk_visual3}
\end{figure}

\begin{table*}[htbp]
\small
\begin{center}
    \caption{Classification Results on the PU Dataset.}
    \label{tab:pu_classification}
    \begin{adjustbox}{width=1.0\textwidth}
\begin{tabular}{
>{\centering\arraybackslash}p{2.0cm}|
>{\centering\arraybackslash}p{1.6cm}|
*{9}{>{\centering\arraybackslash}p{1.5cm}|}
>{\centering\arraybackslash}p{1.5cm}
}

\toprule 
Method & Metric & Jpeg & Zero-Mean & Additive & Poisson & Salt\&Pepper & Stripe & Deadline & Blur & Fog & Avg.\\  
\midrule \midrule 

\multirow{3}{*}{Tent}   
&OA (\%) & $63.83$ & $44.39$ & $69.97$ & $52.59$ & $55.15$ & $61.63$ & $46.33$ & $73.50$ & $59.39$ & $58.53$\\
&AA (\%) & $53.23$ & $32.92$ & $57.30$ & $43.35$ & $44.81$ & $56.04$ & $41.51$ & $60.39$ & $37.20$ & $47.42$\\
&Kappa×100  & $50.62$ & $24.74$ & $56.91$ & $36.89$ & $38.60$ & $49.34$ & $30.49$ & $62.16$ & $40.87$ & $43.96$\\
\midrule
\multirow{3}{*}{BNAdapt}   
&OA (\%) & $50.27$ & $39.22$ & $46.13$ & $51.46$ & $44.09$ & $48.60$ & $36.43$ & $72.73$ & $71.22$ & $51.35$\\
&AA (\%) & $33.23$ & $20.30$ & $28.66$ & $31.94$ & $25.45$ & $32.72$ & $21.58$ & $59.65$ & $53.97$ & $34.61$\\
&Kappa×100  & $31.08$ & $15.65$ & $26.35$ & $33.70$ & $23.32$ & $30.59$ & $16.71$ & $61.36$ & $59.47$ & $33.80$\\
\midrule
\multirow{3}{*}{CoTTA}   
&OA (\%) & $79.23$ & $21.89$ & $81.38$ & $36.24$ & $33.24$ & $70.48$ & $46.83$ & $80.63$ & $8.39$ & $51.59$\\
&AA (\%) & $78.94$ & $44.23$ & $81.85$ & $59.60$ & $56.03$ & $82.39$ & $59.29$ & $74.61$ & $19.89$ & $61.65$\\
&Kappa×100  & $73.03$ & $15.64$ & $75.34$ & $29.41$ & $26.55$ & $63.77$ & $37.23$ & $72.82$ & $4.78$  & $44.29$\\
\midrule
\multirow{3}{*}{CFA}   
&OA (\%) & $65.26$ & $9.18$ & $53.67$ & $12.17$ & $45.77$ & $38.47$ & $33.22$ & $61.04$ & $\textbf{93.54}$ & $45.70$\\
&AA (\%) & $55.08$ & $10.27$ & $59.05$ & $11.91$ & $54.77$ & $46.68$ & $39.29$ & $70.92$ & $\textbf{91.40}$ & $48.82$\\
&Kappa×100  & $54.60$ & $-5.71$ & $45.09$ & $-3.07$ & $37.79$ & $29.12$ & $21.54$ & $53.40$ & $\textbf{91.43}$ & $36.13$\\
\midrule
\multirow{3}{*}{SAR}   
&OA (\%) & $52.16$ & $23.93$ & $52.87$ & $24.89$ & $51.48$ & $45.48$ & $38.28$ & $52.21$ & $72.36$ & $45.41$\\
&AA (\%) & $42.58$ & $12.55$ & $42.61$ & $11.23$ & $41.93$ & $37.79$ & $31.47$ & $48.72$ & $69.81$ & $37.52$\\
&Kappa×100  & $34.57$ & $-7.98$ & $34.56$ & $-5.45$ & $32.81$ & $25.35$ & $18.77$ & $35.90$ & $62.25$ & $25.31$\\
\midrule
\multirow{3}{*}{SOTTA}   
&OA (\%) & $68.44$ & $15.35$ & $54.71$ & $20.40$ & $47.39$ & $43.25$ & $37.31$ & $59.45$ & $89.12$ & $48.16$\\
&AA (\%) & $57.01$ & $21.21$ & $62.07$ & $25.41$ & $60.01$ & $53.36$ & $41.16$ & $70.62$ & $86.85$ & $53.52$\\
&Kappa×100  & $57.67$ & $1.76$  & $46.28$ & $5.68$  & $39.54$ & $33.30$ & $23.98$ & $51.24$ & $85.72$ & $38.68$\\
\midrule
\multirow{3}{*}{Ours}
& OA(\%) & $\textbf{88.32}$ & $\textbf{75.71}$ & $\textbf{89.15}$ & $\textbf{81.52}$ & $\textbf{85.32}$ & $\textbf{94.43}$ & $\textbf{73.24}$ & $\textbf{81.92}$ & $72.69$ & $\textbf{82.48}$\\
&AA(\%) & $\textbf{81.13}$ & $\textbf{60.67}$ & $\textbf{83.83}$ & $\textbf{67.01}$ & $\textbf{76.42}$ & $\textbf{89.98}$ & $\textbf{71.11}$ & $\textbf{77.30}$ & $54.85$ & $\textbf{73.59}$\\
&Kappa$\times$100 & $\textbf{84.07}$ & $\textbf{67.07}$ & $\textbf{85.15}$ & $\textbf{75.01}$ & $\textbf{80.06}$ & $\textbf{92.63}$ & $\textbf{65.71}$ & $\textbf{74.58}$ & $62.17$ & $\textbf{76.27}$\\
\bottomrule 
\end{tabular} 
\end{adjustbox}
\end{center}
\end{table*}

\begin{table*}[htbp]
\small
\begin{center}
    \caption{Classification Results on the WHLK Dataset.}
    \label{tab:whlk_classification}
    \begin{adjustbox}{width=1.0\textwidth}
\begin{tabular}{
>{\centering\arraybackslash}p{2.0cm}|
>{\centering\arraybackslash}p{1.6cm}|
*{9}{>{\centering\arraybackslash}p{1.5cm}|}
>{\centering\arraybackslash}p{1.5cm}
}
\toprule 
Method & Metric & Jpeg & Zero-Mean & Additive & Poisson & Salt\&Pepper & Stripe & Deadline & Blur & Fog & Avg.\\  
\midrule \midrule 
\multirow{3}{*}{Tent}   
&OA (\%)       & $79.41$ & $71.61$ & $76.00$ & $72.57$ & $61.44$ & $78.82$ & $69.27$ & $78.40$ & $65.63$ & 72.79 \\
&AA (\%)       & $56.32$ & $54.80$ & $57.65$ & $51.00$ & $49.32$ & $60.10$ & $48.45$ & $58.04$ & $48.31$ & 53.89 \\
&Kappa×100    & $72.37$ & $62.09$ & $67.79$ & $63.29$ & $48.35$ & $72.06$ & $59.51$ & $71.22$ & $54.14$ & 63.43 \\
\midrule
\multirow{3}{*}{BNAdapt}   
&OA (\%)       & $41.31$ & $45.86$ & $45.32$ & $53.12$ & $30.16$ & $45.82$ & $40.57$ & $59.14$ & $54.55$ & 46.65 \\
&AA (\%)       & $26.37$ & $27.93$ & $28.37$ & $37.61$ & $20.41$ & $30.64$ & $25.28$ & $41.46$ & $39.97$ & 31.78 \\
&Kappa×100    & $24.05$ & $27.77$ & $27.10$ & $37.70$ & $7.35$  & $29.65$ & $22.81$ & $45.66$ & $39.74$ & 29.87 \\
\midrule
\multirow{3}{*}{CoTTA}   
&OA (\%)       & $72.17$ & $37.98$ & $42.45$ & $47.62$ & $19.49$ & $93.41$ & $\textbf{72.20}$ & $68.13$ & $45.02$ & 55.83 \\
&AA (\%)       & $75.82$ & $52.38$ & $57.89$ & $54.29$ & $36.57$ & $94.67$ & $71.40$ & $77.92$ & $50.31$ & 63.47 \\
&Kappa×100    & $66.31$ & $30.62$ & $34.33$ & $39.23$ & $14.74$ & $91.50$ & $\textbf{65.95}$ & $62.06$ & $37.02$ & 49.64 \\
\midrule
\multirow{3}{*}{CFA}   
&OA (\%)       & $67.03$ & $58.10$ & $53.64$ & $71.45$ & $50.04$ & $45.60$ & $42.08$ & $71.82$ & $69.23$ & 58.89 \\
&AA (\%)       & $49.28$ & $44.00$ & $38.62$ & $62.81$ & $41.32$ & $33.03$ & $25.59$ & $67.18$ & $63.83$ & 47.52 \\
&Kappa×100    & $59.17$ & $50.05$ & $45.15$ & $64.88$ & $41.60$ & $33.03$ & $29.91$ & $65.48$ & $62.60$ & 50.65 \\
\midrule
\multirow{3}{*}{SAR}   
&OA (\%)       & $53.19$ & $55.90$ & $53.34$ & $55.10$ & $47.33$ & $44.61$ & $43.27$ & $56.29$ & $57.15$ & 51.80 \\
&AA (\%)       & $27.57$ & $33.20$ & $30.37$ & $30.58$ & $27.07$ & $23.61$ & $19.68$ & $42.38$ & $43.15$ & 30.63 \\
&Kappa×100    & $36.44$ & $39.56$ & $36.09$ & $37.19$ & $27.64$ & $24.14$ & $22.64$ & $41.48$ & $33.97$ & 33.46 \\
\midrule
\multirow{3}{*}{SOTTA}   
&OA (\%)       & $76.10$ & $85.53$ & $83.73$ & $\textbf{87.64}$ & $82.76$ & $60.15$ & $52.76$ & $83.34$ & $82.84$ & 77.99 \\
&AA (\%)       & $42.44$ & $52.34$ & $49.89$ & $57.20$ & $45.57$ & $29.06$ & $24.87$ & $72.01$ & $\textbf{69.23}$ & 49.51 \\
&Kappa×100    & $68.22$ & $80.37$ & $77.98$ & $\textbf{83.37}$ & $76.70$ & $45.30$ & $38.13$ & $78.42$ & $77.98$ & 69.72 \\

\midrule
\multirow{3}{*}{Ours}   
&OA(\%) & \textbf{87.72} &\textbf{ 97.08} & \textbf{98.70} & 60.58 & \textbf{97.25} & \textbf{99.10} & 65.16 & \textbf{92.16} & \textbf{87.86} &\textbf{87.40}\\
&AA(\%) & \textbf{93.29} & \textbf{90.01} & \textbf{95.81} & \textbf{74.44} & \textbf{90.38} & \textbf{98.36} & \textbf{76.66} & \textbf{85.30} & 64.69 &\textbf{86.66}\\
&Kappa×100 & \textbf{85.32} & \textbf{96.15} & \textbf{98.29} & 54.11 & \textbf{96.38} & \textbf{98.82} & 58.98 & \textbf{89.83} & \textbf{83.27} &\textbf{84.68}\\
\bottomrule 
\end{tabular} 
\end{adjustbox}
\end{center}
\end{table*}

\begin{figure*}[htbp]
\centering
\includegraphics[width=1.0\textwidth]{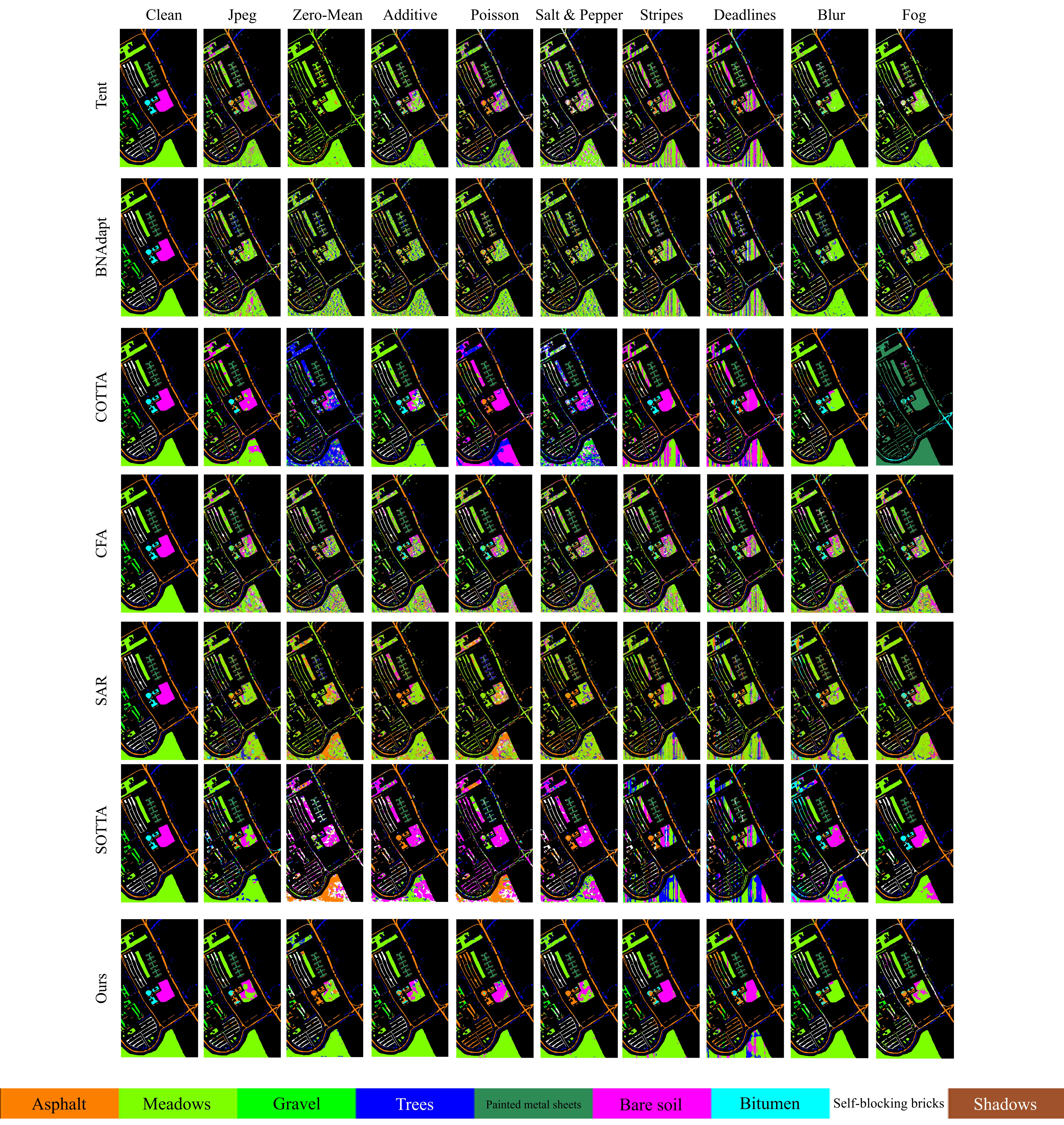}
\caption{Visual classification results on the PU dataset under different degradation types after applying different TTA method. Each row corresponds to a method, and each column represents a specific degradation.}
\label{fig:pu_visual2}
\end{figure*}

\begin{figure*}[htbp]
\centering
\includegraphics[width=1.0\textwidth]{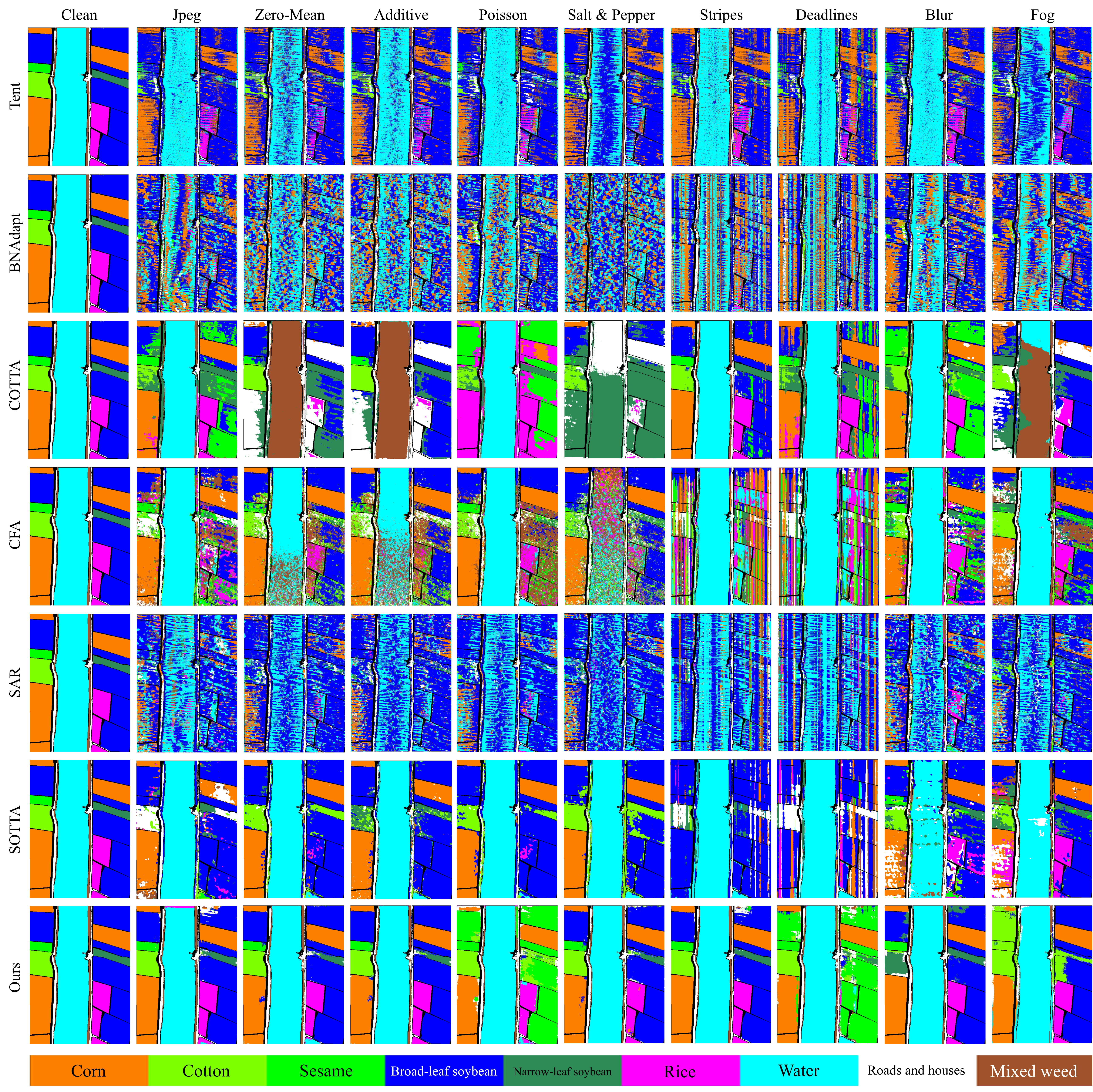}
\caption{Visual classification results on the WHLK dataset under different degradation types after applying different TTA method. Each row corresponds to a method, and each column represents a specific degradation.}
\label{fig:whlk_visual2}
\end{figure*}

\begin{table*}[htbp]
\small
\begin{center}
    \caption{Ablation Study on MRF, LSL, and CELA Modules on WHLK Dataset.}
    \label{tab:ablation}
    \begin{adjustbox}{width=1.0\textwidth}
    
\begin{tabular}{
>{\centering\arraybackslash}p{2.0cm}|
>{\centering\arraybackslash}p{1.6cm}|
*{9}{>{\centering\arraybackslash}p{1.5cm}|}
>{\centering\arraybackslash}p{1.5cm}
}
\toprule Method& Metric &Clean &Jpeg& Zero-Mean& Additive& Poisson& Salt\&Pepper& Stripe& Deadline& Blur& Fog\\   
\midrule \midrule 
\multirow{3}{*}{Ours}   
&OA (\%) &99.8 & 87.72 & 97.08 & 98.70 & 60.58 & 97.25 & 99.10 & 65.16 & 92.16 & 87.86 \\
&AA (\%) &99.17 & 93.29 & 90.01 & 95.81 & 74.44 & 90.38 & 98.36 & 76.66 & 85.30 & 64.69 \\
&Kappa×100 &99.73 & 85.32 & 96.15 & 98.29 & 54.11 & 96.38 & 98.82 & 58.98 & 89.83 & 83.27 \\
\midrule
\multirow{3}{*}{w/o MRF} 
&OA (\%) & 99.85 & 63.84 & 52.07 & 60.72 & 59.15 & 64.00 & 98.80 & 66.47 & 84.59 & 29.52 \\
&AA (\%) & 99.57 & 75.95 & 58.67 & 66.96 & 78.80 & 68.94 & 99.03 & 81.37 & 80.03 & 35.48 \\
&Kappa×100 & 99.80 & 54.94 & 46.24 & 54.31 & 52.72 & 58.68 & 98.43 & 60.54 & 80.71 & 22.82 \\
\midrule
\multirow{3}{*}{w/o LSL}
&OA (\%) & 99.66 & 69.82 & 71.97 & 93.93 & 61.68 & 78.21 & 96.44 & 64.58 & 86.45 & 54.40 \\
&AA (\%) & 98.98 & 77.72 & 73.68 & 92.89 & 74.98 & 77.23 & 96.95 & 77.14 & 80.34 & 49.79 \\
&Kappa×100 & 99.55 & 62.27 & 66.39 & 91.94 & 55.34 & 73.56 & 95.41 & 58.39 & 82.77 & 46.58 \\
\midrule
\multirow{3}{*}{w/o CELA}
&OA (\%) &99.8 & 85.06 & 86.25 & 92.14 & 56.32 & 87.76 & 95.62 & 69.09 & 85.96 & 64.09 \\
&AA (\%) &99.17 & 92.11 & 84.07 & 90.20 & 70.43 & 89.08 & 96.83 & 80.65 & 81.21 & 63.18 \\
&Kappa×100 &99.73 & 81.24 & 82.77 & 89.90 & 49.47 & 84.50 & 94.33 & 63.14 & 82.21 & 57.62 \\
\bottomrule 
\end{tabular} 
\end{adjustbox}
\end{center}
\end{table*}

\subsubsection{Comparison of Classification Models under Multi-Degradation}
To evaluate the robustness and generalization ability of the proposed classification model SSTC, we compare it against several representative image classification models, including SSRN~\cite{8061020}, SSFTT~\cite{9684381}, CASST~\cite{9874815}, and MASSFormer~\cite{10506482}. All models are trained under identical conditions on the clean data and evaluated under various degradation scenarios simulated on the PU and WHLK datasets. This comparison aims to demonstrate the intrinsic robustness of SSTC in challenging environments, which is essential for the success of subsequent TTA.

Fig.~\ref{fig:pu_visual1} presents the classification accuracy (OA) of different models under various degradation types on the PU dataset. The solid lines represent the performance of each model across different degradations, while the dashed lines indicate the average OA of each model over all degradation scenarios. As shown in the figure, the proposed SSTC consistently achieves higher accuracy across most degradation types compared to existing models, including SSFTT, SSRN, CASST, and MASSFormer. Notably, SSTC maintains superior performance under severe degradations such as Poisson noise, Gaussian noise (ZM), and stripes. In addition, SSTC exhibits the highest average OA (as indicated by the topmost blue dashed line), demonstrating its strong robustness and stability under diverse degradation conditions. These results verify that SSTC provides a more reliable foundation for subsequent TTA.

Fig.~\ref{fig:whlk_visual1} illustrates the classification accuracy (OA) of different models under various degradation types on the WHLK dataset. Compared to other baselines, the proposed SSTC exhibits superior robustness across most degradation types, including challenging scenarios such as ZM Gaussian noise and Salt and pepper noise. While SSRN and SSFTT suffer from significant performance drops under certain corruptions, SSTC maintains high accuracy and stable trends across the board. Although the average OA of SSTC is slightly lower than that of MASSFormer, it remains very close, as indicated by the nearly overlapping dashed lines. This demonstrates that SSTC achieves comparable overall performance while offering greater consistency across degradations. These results confirm that SSTC provides a reliable and robust foundation for downstream TTA methods.

Above results show that SSTC  outperforms the compared models across most degradation types on both datasets, achieving higher average OA scores. These results indicate that SSTC is not only more robust and stable under distribution shifts, but also provides a solid foundation for TTA. We next compare our complete HyperTTA framework with existing TTA methods under the same degradation conditions.

\subsubsection{Comparison of TTA Methods}
To assess the effectiveness of the proposed HyperTTA framework, we compare it with several state-of-the-art TTA methods, including Tent~\cite{wang2020tent}, CoTTA~\cite{wang2022continual}, BNAdapt~\cite{LI2018109}, CFA~\cite{kojima2022robustvit}, SAR~\cite{niu2023towards}, and SoTTA~\cite{gong2023sotta}. For a fair comparison, Tent, CoTTA, and BNAdapt are implemented using SSRN as the backbone, while CFA, SAR, and SoTTA adopt SSFTT as their base model. Our HyperTTA is built upon the proposed SSTC backbone. All methods are evaluated on the same degraded test sets of the PU and WHLK datasets.

Table~\ref{tab:pu_classification} reports the classification performance (OA, AA, Kappa) of different TTA methods on the PU dataset under multi-degradations, and the visualized OA performance is shown in Fig.~\ref{fig:PU_visual3}. Compared with all baselines, the proposed HyperTTA achieves higher classification accuracy across most degradation types. Specifically, it outperforms other methods under challenging conditions such as additive noise (89.15\%), Poisson noise (81.52\%), and deadlines (73.24\%), while maintaining strong performance on milder corruptions such as JPEG and S\&P noise. Moreover, HyperTTA achieves the highest average performance, with an OA of 82.48\%, AA of 73.59\%, and Kappa of 76.27\%, demonstrating superior robustness and generalization capability. Fig.~\ref{fig:pu_visual2} further visualizes the classification maps produced by different TTA methods under various degradations. It can be observed that our method preserves finer spatial structures and generates fewer artifacts, particularly under heavy noise conditions (e.g., Poisson, stripes, ZM Gaussian), while other methods often produce fragmented or noisy predictions. 

Table~\ref{tab:whlk_classification} shows the classification results of various TTA methods on the WHLK dataset under multi-degradations, and the visualized OA performance is shown in Fig.~\ref{fig:whlk_visual3}. Overall, our HyperTTA method outperforms all baselines by a significant margin in terms of OA, AA, and Kappa metrics. In particular, it achieves an average OA of 87.40\%, AA of 86.66\%, and Kappa of 84.68\%, which are markedly higher than the second-best method, SoTTA (OA: 77.99\%). HyperTTA maintains high performance across nearly all degradation types, especially on challenging conditions such as Zero-Mean Gaussian (97.08\%), Salt\&Pepper noise (97.25\%), and fog (88.86\%). The classification maps in Fig.~\ref{fig:whlk_visual2} further illustrate the effectiveness of each TTA method. Our method consistently produces clean, spatially coherent, and visually accurate results under all degradations. In contrast, the other methods exhibit severe noise, class confusion, and boundary distortion, particularly under additive, Poisson, and stripe corruptions. 

Above results demonstrate that HyperTTA achieves the highest overall accuracy across most degradation types and datasets, validating the effectiveness of both our robust backbone and the confidence-aware entropy minimization strategy. In particular, HyperTTA exhibits superior performance stability under severe degradations, highlighting its practical potential in real-world hyperspectral applications.

\subsection{Ablation Study}
Table~\ref{tab:ablation} presents the ablation study results on the three key components of the proposed framework, including the Multi-level Receptive Field (MRF) module, the Label Smoothing Loss (LSL) module, and the Confidence-aware Entropy-minimized LayerNorm Adapter (CELA) module. Each component is ablated individually to assess its contribution under various degradation types. This experiment is conducted on the WHLK dataset to ensure a comprehensive evaluation under complex and realistic conditions.

The complete model ("Ours") consistently achieves the highest classification accuracy across most degradation types. Notably, it obtains 87.72\% OA on JPEG noise, 97.08\% on zero-mean Gaussian noise, and 87.86\% on fog degradation, indicating strong robustness to diverse noise patterns. Removing the MRF module causes a significant performance drop under complex degradations such as fog (from 87.86\% to 29.52\%) and zero-mean Gaussian noise (from 97.08\% to 52.07\%), suggesting its importance for modeling multi-scale spatial consistency. Excluding the LSL results in a moderate but consistent decline across most conditions, especially under blur and fog. The absence of the CELA module leads to clear degradation in performance under several noise types, such as blur (OA drops from 92.16\% to 85.96\%) and fog (from 87.86\% to 64.09\%).

Overall, these results demonstrate that each component contributes to the overall robustness of the proposed framework, and their combination leads to significantly improved performance across diverse degradation scenarios.

\section{Conclusion}
In this paper, we investigate the robustness of HSI classification under a wide range of degradations and propose a framework named HyperTTA (Test-Time Adaptable Transformer for Hyperspectral Degradation) to enhance model robustness against diverse degradation conditions. Specifically, we first construct a comprehensive multi-degradation hyperspectral dataset that simulates nine representative types of corruptions, enabling systematic evaluation of model performance under challenging scenarios. Based on this dataset, we develop a spectral–spatial Transformer-based classification framework that incorporates a multi-level receptive field mechanism and label smoothing regularization to jointly enhance spatial representation and generalization capability. Furthermore, we introduce a lightweight TTA strategy that updates only the affine parameters of LayerNorm layers via entropy minimization. To improve the stability and reliability of the adaptation process, this update is performed selectively on high-confidence unlabeled target samples. This confidence-aware strategy enables robust, source-free adaptation to distribution shifts without requiring access to source data or target annotations. Experimental results on PU and WHLK datasets confirm that HyperTTA achieves superior performance under various degradation conditions, demonstrating its robustness and generalization capability.

While HyperTTA demonstrates promising performance, it has several limitations. The current TTA strategy relies on clean source model weights, which may limit its applicability to models under extreme distribution shifts. Additionally, the degradation simulations are based on synthetic corruptions and do not yet include real-world degraded hyperspectral data. In future work, we will enhance the realism of the dataset by incorporating physically acquired degradations (e.g., fog, sensor failures), and extend the framework to dynamic scenarios such as temporally evolving scenes and online adaptation in streaming settings.


%

\FloatBarrier

\ifCLASSOPTIONcaptionsoff
  \newpage
\fi



%
\bibliographystyle{IEEEtran}


%





\FloatBarrier
\end{document}